\documentclass[sn-mathphys-num]{sn-jnl}

 \usepackage[table,xcdraw]{xcolor}
\usepackage{graphicx}%
\usepackage{multirow}%
\usepackage{amsmath,amssymb,amsfonts}%
\usepackage{amsthm}%
\usepackage{mathrsfs}%
\usepackage[title]{appendix}%
\usepackage{textcomp}%
\usepackage{manyfoot}%
\usepackage{booktabs}%
\usepackage{algorithm}%
\usepackage{algorithmicx}%
\usepackage{algpseudocode}%
\usepackage{listings}%
\usepackage{booktabs}

\usepackage{caption}
\usepackage{graphicx}
\usepackage{float} 
\usepackage{subcaption}
\usepackage{tabularx}
\usepackage{hhline}
\usepackage[authoryear, round]{natbib}

\usepackage{longtable}
\usepackage{geometry}

\usepackage{array}
\usepackage{makecell}
\usepackage{tikz}
\usetikzlibrary{mindmap, shapes.geometric, arrows.meta}
\usepackage{amsthm}

\usepackage{tikz}
\usetikzlibrary{mindmap}
\definecolor{col1}{HTML}{DDEFF4}
\definecolor{col2}{HTML}{FDEBDD}
\definecolor{col3}{HTML}{EAF6E1}
\definecolor{col4}{HTML}{E8E5FC}
\definecolor{col5}{HTML}{FEF9CD}

\usepackage{geometry}
\usepackage{longtable}

\begin{document}

\title[Article Title]{Deep Learning for Time Series Forecasting: A Survey}








\author[1]{\fnm{Xiangjie} \sur{Kong}}\email{xjkong@ieee.org}

\author[1]{\fnm{Zhenghao} \sur{Chen}}\email{202006010504@zjut.edu.cn}

\author[1]{\fnm{Weiyao} \sur{Liu}}\email{211123120037@zjut.edu.cn}
\author[1]{\fnm{Kaili} \sur{Ning}}\email{211124120018@zjut.edu.cn}

\author[1]{\fnm{Lechao} \sur{Zhang}}\email{202003151126@zjut.edu.cn}
\author[1]{\fnm{Syauqie} \sur{Muhammad Marier}}\email{syauqie.mm@gmail.com}
\author[1]{\fnm{Yichen} \sur{Liu}}\email{liuien@outlook.com}
\author[1]{\fnm{Yuhao} \sur{Chen}}\email{yuhaochen4859@gmail.com}
\author*[2]{\fnm{Feng} \sur{Xia}}\email{f.xia@ieee.org}

\affil[1]{\orgdiv{College of Computer Science and Technology}, \orgname{Zhejiang University of Technology}, \orgaddress{ \city{Hangzhou}, \postcode{310023}, \country{China}}}

\affil*[2]{\orgdiv{School of Computing Technologies}, \orgname{RMIT University}, \orgaddress{ \city{Melbourne}, \postcode{3000},  \country{Australia}}}


\abstract{Time series forecasting (TSF) has long been a crucial task in both industry and daily life. Most classical statistical models may have certain limitations when applied to practical scenarios in fields such as energy, healthcare, traffic, meteorology, and economics, especially when high accuracy is required. With the continuous development of deep learning, numerous new models have emerged in the field of time series forecasting in recent years. However, existing surveys have not provided a unified summary of the wide range of model architectures in this field, nor have they given detailed summaries of works in feature extraction and datasets. To address this gap, in this review, we comprehensively study the previous works and summarize the general paradigms of Deep Time Series Forecasting (DTSF) in terms of model architectures. Besides, we take an innovative approach by focusing on the composition of time series and systematically explain important feature extraction methods. Additionally, we provide an overall compilation of datasets from various domains in existing works. Finally, we systematically emphasize the significant challenges faced and future research directions in this field.}




\maketitle

\section{Introduction}\label{sec1}

Time series are pervasive in various facets of our manufacture and life, serving as a primary dimension to record historical events. Forecasting, a critical task, leverages historical information within sequences to infer the future  \citep{hyndman2018forecasting,petropoulos2022forecasting}. It finds extensive applications in various domains closely intertwined with our lives, including energy production and consumption  \citep{deb2017review, li2019text, saxena2019hybrid, rajagukguk2020review, zhao2016novel, toubeau2018deep}, meteorological variations  \citep{mudelsee2019trend,zaini2022systematic}, finance, stock markets, and econometrics   \citep{sezer2020financial, chen2015hybrid, callot2017modeling, andersen2005volatility, luo2018neural, kalra2024efficient, singh2023analysis} sales and demand  \citep{bose2017probabilistic,bandara2019sales} urban traffic flows  \citep{tedjopurnomo2020survey, lv2014traffic, li2017diffusion}, and welfare-related healthcare conditions  \citep{piccialli2021artificial, kaushik2020ai, topol2019high, mishra2024multivariate}.

Machine learning, data science, and other research groups employing operations research and statistical methods have extensively explored time series forecasting \citep{fuller2009introduction, faloutsos2018forecasting, faloutsos2019classical, faloutsos2019forecasting}. Statistical models typically consider non-stationarity, linear relationships, and specific probability distributions to infer future trends based on the statistical properties of historical data such as mean, variance, and autocorrelation. On the other hand, machine learning models learn patterns and rules from the data. With the emergence \citep{rosenblatt1957perceptron} and rapid development of deep learning \citep{goodfellow2016deep,lecun2015deep}, an increasing number of neural network models are being applied to time series forecasting. In contrast to the first two approaches that rely on domain-specific knowledge or meaningful feature engineering, deep learning autonomously extracts intricate time features and patterns from complex data. This capability enables the capture of long-term dependencies and complex relationships, ultimately enhancing prediction accuracy. In this article, we will refer to works on Deep Learning for Time Series Forecasting as DTSF works, and Time Series Forecasting will be abbreviated as TSF.

In recent years, deep learning methods have continuously advanced and innovated in time series forecasting (TSF) across various domains \citep{salinas2020deepar, xu2016quantile, lai2018modeling, bandara2020lstm, oord2016wavenet, rasul2021autoregressive, lim2021temporal}. However, current research efforts primarily focus on key TSF concepts and fundamental model components, while lacking a high-level categorization of deep learning-based DTSF model structures, comprehensive summaries of recent developments, and in-depth analyses of future prospects and challenges. This article aims to address these gaps by drawing on the latest research. The main contributions of this work are as follows:

\begin{itemize}
\item  \textbf{Dynamic and systematic taxonomy.} We propose a novel dynamic classification method designed to categorize deep learning models for time series forecasting in a systematic manner. Our survey classifies and summarizes these models from the perspective of their architectural structure. To the best of our knowledge, this represents the first dynamic classification of deep learning model architectures for time series forecasting.
\end{itemize}

\begin{itemize}
\item \textbf{Comprehensive review of data feature enhancement.} We analyze and summarize feature enhancement methods for time series data, including dimensional decomposition, time-frequency transformation, pre-training, and patch-based segmentation. Our analysis begins with the composition of complex, high-dimensional data features, aiming to reveal the latent learning potential within time series data.
\end{itemize}

\begin{itemize}
\item \textbf{Summary of challenges and future directions.} This survey summarizes major TSF datasets from recent years, discusses key challenges, and highlights promising future research directions to advance the field.
\end{itemize}

\begin{figure*}[!t]
    \centering
    \setlength{\abovecaptionskip}{4mm}
    \captionsetup{labelsep=none} 
    \includegraphics[width=\linewidth,height=\textheight,keepaspectratio]{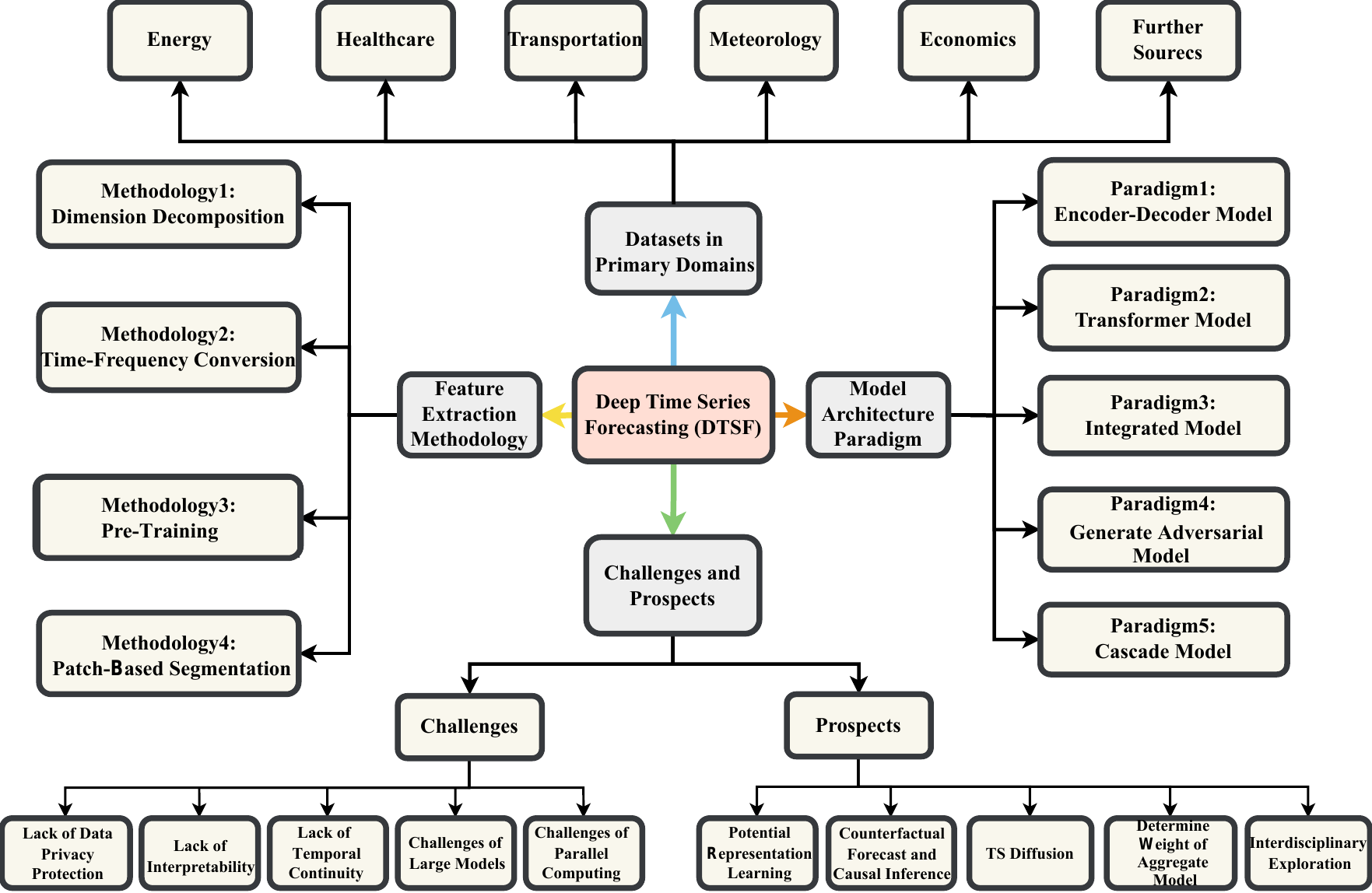}
  \captionsetup{name={Fig.},justification=raggedright} 
    \caption{\hspace{1em}The outline of this article}
    \vspace{-0.1in}
    \label{fig1}
    
\end{figure*}

The remaining content is organized as follows. Section \ref{sec2} introduces the fundamental aspects of TSF, encompassing the definition and composition of time series, forecasting tasks, statistical models, and existing problems. Section \ref{sec3}, a pivotal component of this paper, mainly delineates the overarching structural paradigm of DTSF models. Section \ref{sec4} outlines the prevalent paradigms for extracting and learning features from time series data, constituting the second major focus. Section \ref{sec6} is another key focus of this paper. We not only highlight the limitations and challenges within the current achievements in DTSF research but also elucidate prospective avenues for future exploration. Finally, we conclude this survey in Section \ref{sec7}. In Appendix \ref{sec5}, an exhaustive account of TSF datasets across various domains is presented.  Figure \ref{fig1} shows an outline of the entire paper.

\section{Time Series Forecasting}\label{sec2}

Time series represents a continuous collection of data points recorded at regular or irregular time intervals, offering a chronological record of observed phenomena such as vital signs, sales trends, stock market prices, weather changes, and more. The nature of these observations can encompass numerical values, labels, etc. Moreover, time series can be either discrete or continuous \citep{hamilton2020time}. It is commonly employed for the analysis and prediction of trends and patterns \citep{montgomery2015introduction} that evolve over time.

TSF is the process of forecasting future values based on the inherent properties and characteristic patterns found in historical data. These properties and intrinsic patterns may provide valuable insights into describing future occurrences. Discovering potential features within time series data based on the similarity of statistical characteristics between adjacent data points or time steps is crucial for building a strong foundation for designing prediction models and achieving improved results.

In this section, we will begin with the definition of time series and explain the concept of TSF. Furthermore, we will introduce classical methods based on mathematical statistics. Lastly, we will analyze the factors contributing to lower prediction accuracy to provide researchers new to this field with a preliminary understanding.

\subsection{Time Series Definition}
In this survey, we consider time series as observation sequences recorded in chronological order, which may have fixed or variable time intervals between observations. Let $t$ denote the time of observation, and $\mathbf{y}_t$ represents the time series, corresponding to a stochastic process composed of random variables observed over time. In most cases, $t \in \mathbb{Z}$, where $\mathbb{Z} = (0, \pm 1, \pm 2, \ldots)$ represents the set of positive and negative integers \citep{fuller2009introduction}. When only a limited amount of data is available, a time series can be represented as $(\mathrm{y}_1, \mathrm{y}_2, \mathrm{y}_3, \ldots)$. Let $\mathcal{Y} = \{\mathbf{y}_{i,1:T_i}\}_{i=1}^N$ denote the collection of N univariate time series, where $\mathbf{y}_{i,1:T_i} = (y_{i,1}, \ldots, y_{i,T_i})$, and $y_{i,t}$ represents the values of $t$ for the i-th time series. $\mathbf{Y}_{t_1:t_2}$ is the collection of values for all $N$ time series within the time interval $[t_1, t_2]$.

Time series data differs from other forms of data since it is prevalent in all major fields and is significant as one of the aspects that make up our reality. It has a wide range of attributes and characteristics. First of all, time series data are usually noisy and high-dimensional. Techniques such as dimensionality reduction, wavelet analysis, or filtering can be used to eliminate some noise and reduce dimensionality \citep{zebari2020comprehensive}. Secondly, the sample time interval has an impact on it. Due to its inherent instability in reality, the distribution of time series obtained at different sampling frequencies does not have a uniform probability distribution \citep{yalavarthi2024grafiti}. Finally, if time series data is viewed as an information network, each time point can be considered a node, with the relationships between nodes evolving over time. Similar to most real-world networks, this data is inherently heterogeneous and dynamic \citep{peng2021lime}, which presents significant challenges for the modeling and analysis of spatio-temporal data. It is worth noting that the representation of time series data is crucial for relevant features extraction and dimensionality reduction. The success or failure of model design and application is closely tied to this representation.

\subsection{Forecasting Task}
TSF is a process of predicting future data based on historical observations, widely applied in various domains such as energy, finance, and meteorology to anticipate future trends. The task of TSF can be categorized into short-term and long-term forecasting based on the prediction horizon, which is determined by specific application requirements and domain characteristics. Short-term forecasting typically involves shorter time spans, often ranging from hours to weeks, emphasizing high prediction accuracy and is suitable for tasks demanding precision. In contrast, long-term forecasting spans longer periods, including months, years, or even longer durations, and addresses challenges related to long-term trends and seasonal variations that can significantly impact prediction accuracy. The distinction between these two types of forecasting lies in their specific emphasis. Short-term forecasting prioritizes precision and relies mainly on extrapolating data, suitable for scenarios where fluctuations within relatively short periods are critical for prediction outcomes. Conversely, long-term forecasting requires consideration of long-term trends and seasonal influences, making it more complex and necessitating additional factors such as extra assumptions and supplemental external data, which may affect its accuracy. Therefore, the role of external factors is particularly important in long-term forecasting, as they help the forecasting model better capture long-term trends, cyclical fluctuations, and other macro-level changes. For example, external factors such as weather, holidays, economic indicators, and road network information often have a significant impact on the trends and seasonal variations in time series data. Currently, many researchers have incorporated these external factors into forecasting models to improve the accuracy of predictions. Common approaches to handling external influences include incorporating external data as additional features into the model, using multi-task learning with external data \citep{ruder2017overview}, and introducing exogenous variables into classical time series models. Deep learning methods, such as LSTM, GRU, and attention mechanisms, also enhance model performance by considering external factors \citep{qin2017dual}. Additionally, seasonal adjustment, periodic modeling, and the integration of road network knowledge are effective methods for addressing external influences. For instance, MultiSPANS \citep{zou2024multispans} uses a structural entropy minimization algorithm to generate optimal road network hierarchies, considering complex multi-distance dependencies in the road network for prediction; \citep{kong2024exploring}, in summarizing forecasting tasks, constructed a new bus station distance network to account for the relationships between external bus stations.

On the other hand, in addition to being categorized as Univariate \citep{zhang1998forecasting, januschowski2020criteria, montero2021principles, semenoglou2021investigating} and Multivariate \citep{lutkepohl2005vector,kolassa2020best} forecasting based on whether multiple variables are considered, TSF can also be distinguished by the distinction between global and local models. Univariate forecasting involves tasks where only one variable is considered during the forecasting process, primarily focusing on predicting the future values of a single variable. Multivariate forecasting, on the other hand, entails the simultaneous prediction of multiple correlated variables, considering the interdependencies among various variables and forecasting their future values. When discussing univariate and multivariate forecasting, it's essential to consider the distinction between global and local models, which impacts the modeling approach and the interpretation of results. Global models consider all variables across the entire time series dataset, while local models focus on subsets of the data, such as specific segments or windows, affecting how dependencies within the data are captured and predictions are made.

In summary, the categorization and focus of forecasting tasks depend on the application context and requirements. For instance, in the financial domain, short-term forecasting may involve predicting stock price fluctuations within minutes or hours, while long-term forecasting could encompass forecasts over several weeks or months. Similarly, in meteorology, short-term forecasting might entail predicting weather conditions within a few hours, while long-term forecasting may involve predictions spanning days or weeks. For univariate forecasting, the focus could be on forecasting the sales volume of a particular product or the price of a specific stock. On the other hand, multivariate forecasting might simultaneously predict the sales volumes of multiple products or the interrelationships within various financial markets.

In the following subsections, we will introduce statistical forecasting models and highlight their limitations, emphasizing the challenges posed by traditional TSF methods. Subsequently, we will delve into the development of deep learning forecasting models and methods.

\subsection{Statistical Forecasting Model}
The development history of statistical forecasting models can be traced back to the early 20th century. Equations 1 and 2 illustrate how the first statistical forecasting methods, such as Moving Averages (MA) \citep{box2015time, hipel1994time, cochrane1997time} and simple Exponential Smoothing (ES) \citep{gardner1985exponential}, were based on time series.

\begin{equation}
\label{deqn_ex1}
MA_t(n)=\frac1n\sum_{i=t-n+1}^tx_i
\end{equation}
where ${n}$ is the window size, and MA represents the moving average at time ${t}$. 

\begin{equation}
\label{deqn_ex2}
ES_{t+1}=\alpha \cdot x_t+(1-\alpha )\cdot ES_t
\end{equation}
where $ES_{t+1}$ represents the predicted trend, \(\alpha\) is the smoothing coefficient, and $ES_t$ is the value predicted at the previous time step. Moving average smooths data by calculating the average of observed values over a certain period of time, while exponential smoothing assigns higher weights to more recent observations to reflect the trend of the data.

Subsequently, the autoregressive (AR) \citep{box2015time, hipel1994time, lee1994univariate} and Moving Average (MA) models (represented by Equations 3 and 4, respectively) were introduced as two important concepts, leading to the development of the Autoregressive Moving Average Model \citep{box2015time, hipel1994time, adhikari2013introductory} (ARMA, as shown in equation 5). These models aim to accurately capture the auto correlation and averaging properties of time series data.

\begin{equation}
\label{deqn_ex3}
AR\colon Y_t = c + \varphi_1 Y_{t-1} + \varphi_2 Y_{t-2} + \cdots + \varphi_p Y_{t-p} + \xi_t
\end{equation}

\begin{equation}
\label{deqn_ex4}
MA\colon Y_t = \mu + \epsilon_t + \theta_1 \epsilon_{t-1} + \theta_2 \epsilon_{t-2} + \cdots + \theta_q \epsilon_{t-q}
\end{equation}

\begin{equation}
\begin{aligned}
Y_t = & \, c + \varphi_1 Y_{t-1} + \varphi_2 Y_{t-2} + \cdots + \varphi_p Y_{t-p} \\
      & + \theta_1 \epsilon_{t-1} + \theta_2 \epsilon_{t-2} + \cdots + \theta_q \epsilon_{t-q} + \epsilon_t
\end{aligned}
\end{equation}

\noindent where $Y_{t}$ represents the time series data under consideration, $\varphi_{1}$ to $\varphi_{p}$ are parameters of the AR model. These parameters describe the relationship between the current value and values from the past $p$ time points. Similarly, $\theta_{1}$ to $\theta_{q}$ are parameters of the MA model, which describe the relationship between the current value and errors from the past $q$ time points. $\varepsilon_{t}$ represents the error term at time $t$, and $\text{c}$ denotes a constant term.

Specifically, the AR model leverages past time series observations to predict future values, while the MA model relies on the moving average of observations to make these predictions. To address non-stationary time series data, the Autoregressive Integrated Moving Average (ARIMA) model \citep{box2015time, hipel1994time, cochrane1997time,hamzaccebi2008improving,zhang2003time} is introduced. ARIMA is employed to transform non-stationary sequences into stationary ones by means of differencing, thereby reducing or eliminating trends and seasonal variations in the time series. This transformation is represented by Equation (6) as follows:

\begin{equation}
\label{deqn_ex6}
\Delta Y_t=(1-L)^dY_t=\epsilon_t
\end{equation}

\noindent where \(L\) denotes the lag operator, \(d\) represents the differencing order, \(y_t\) signifies the time series, and \(\epsilon_t\) is the error term. This integration of ARIMA helps mitigate non-stationarity, paving the way for more effective TSF.

Machine learning models represented by Random Forests and Decision Trees \citep{rokach2016decision, ali2012random, ho1995random, kontschieder2015deep} offer enhanced flexibility and predictive performance in statistical forecasting \citep{harvey1990forecasting,ahmed2010empirical}. A decision tree comprises a series of decision nodes and leaf nodes, constructed based on the selection of optimal features and splitting criteria to minimize prediction errors or maximize metrics like information gain or Gini index. Each decision node splits based on feature conditions, while each leaf node provides prediction results. Random Forest, on the other hand, makes forecasting by constructing multiple decision trees and combining their forecasting results. It can handle high-dimensional features and large-scale datasets, capturing nonlinear relationships and interactions between features.

However, the development of emerging technologies such as the Internet of Things (IoT) has brought efficiency and convenience to data acquisition, collection, and storage \citep{li2015internet,kong2022edge}. The era of big data has arrived \citep{sagiroglu2013big,fan2014challenges}, with data being generated at an increasing rate. Statistical forecasting models need to better adapt to the demands of processing large-scale and high-dimensional data \citep{che2013big, wu2013data, oussous2018big}. Different industries and domains are also increasingly in need of accurate forecasting models to support decision-making and planning \citep{rodriguez2016general}. Furthermore, more complex relationships among data are encountered in practical applications, requiring more flexible and accurate models to tackle these challenges.

In summary, traditional statistical forecasting models are limited in terms of computational power, prediction accuracy, and length. There are major shortcomings in statistical forecasting methods in handling non-stationarity, nonlinear relationships, noise, and complex dependencies, and their adaptability to long-term dependencies and multi-feature forecasting tasks is also limited. With the continuous development and innovation of deep learning models, these limitations have been overcome, leading to improved predictive performance.

\begin{table*}[h!]
        \tiny
        \captionsetup{labelsep=none} 
	\caption{\hspace{1em}DTSF Model Architecture Paradigm}
  \vspace{0.3mm}
	\label{DTSF} 

\renewcommand{\arraystretch}{1.03}
\hspace{-2cm}
\begin{tabular}{p{1.2cm}|p{3.5cm}|p{1.3cm}|p{0.6cm}|p{3cm}|p{3cm}|p{0.5cm}}
\hline
\cellcolor[HTML]{EDEBEB} 

 Architecture       &\cellcolor[HTML]{EDEBEB}  Model & \cellcolor[HTML]{EDEBEB} Multi/Uni  & \cellcolor[HTML]{EDEBEB} Output     &\cellcolor[HTML]{EDEBEB} Loss & \cellcolor[HTML]{EDEBEB} Metrics     &\cellcolor[HTML]{EDEBEB}  Year   \\ \hline
                                    
                                    & \cellcolor[HTML]{FFCCC9}COST \citep{woo2022cost} &\cellcolor[HTML]{FFCCC9} Multi \& Uni & \cellcolor[HTML]{FFCCC9}Point & \cellcolor[HTML]{FFCCC9}contrastive loss & \cellcolor[HTML]{FFCCC9}MSE, MAE & \cellcolor[HTML]{FFCCC9}2022               \\ \hhline{~------} 
                                    & \cellcolor[HTML]{FFCCC9}TS2Vec \citep{yue2022ts2vec} & \cellcolor[HTML]{FFCCC9}Multi \& Uni & \cellcolor[HTML]{FFCCC9}Point & \cellcolor[HTML]{FFCCC9}contrastive loss & \cellcolor[HTML]{FFCCC9}MSE & \cellcolor[HTML]{FFCCC9}2022               \\ \hhline{~------} 
                                    &\cellcolor[HTML]{FFCCC9} ACT \citep{li2022act} &\cellcolor[HTML]{FFCCC9} Multi \& Uni &\cellcolor[HTML]{FFCCC9} Point & \cellcolor[HTML]{FFCCC9}cross-entropy & \cellcolor[HTML]{FFCCC9}Q50 loss, Q90 loss &\cellcolor[HTML]{FFCCC9} 2022              \\ \hhline{~------}  
                                    & \cellcolor[HTML]{FFCCC9}SimTS \citep{zheng2023simts} &\cellcolor[HTML]{FFCCC9} Multi \& Uni & \cellcolor[HTML]{FFCCC9}Point &\cellcolor[HTML]{FFCCC9} cos-similarity loss, InfoNCE loss &\cellcolor[HTML]{FFCCC9} MAE, MSE &\cellcolor[HTML]{FFCCC9} 2023  \\ \hhline{~------} 
                                    &\cellcolor[HTML]{FFFFE6} DeepTCN \citep{chen2020probabilistic} &\cellcolor[HTML]{FFFFE6} Multi &\cellcolor[HTML]{FFFFE6} Pro & \cellcolor[HTML]{FFFFE6}quantile loss & \cellcolor[HTML]{FFFFE6}NRMSE, SMAPE, MASE & \cellcolor[HTML]{FFFFE6}2020              \\ \hhline{~------} 
                                    
                                    &\cellcolor[HTML]{FFFFE6} STEP \citep{shao2022pre} & \cellcolor[HTML]{FFFFE6}Multi & \cellcolor[HTML]{FFFFE6}Pro &\cellcolor[HTML]{FFFFE6} MAE & \cellcolor[HTML]{FFFFE6}MAE, RMSE, MAPE & \cellcolor[HTML]{FFFFE6}2022            \\ \hhline{~------} 
                                    
                                    & \cellcolor[HTML]{FFFFE6}DCAN \citep{he2022dynamic} & \cellcolor[HTML]{FFFFE6}Multi & \cellcolor[HTML]{FFFFE6}Point & \cellcolor[HTML]{FFFFE6}RMSE & \cellcolor[HTML]{FFFFE6}MAE, RMSE &\cellcolor[HTML]{FFFFE6} 2022               \\ \hhline{~------}  
& \cellcolor[HTML]{FFFFE6}FusFormer \citep{yang2022fusion} & \cellcolor[HTML]{FFFFE6}Multi & \cellcolor[HTML]{FFFFE6}Point & \cellcolor[HTML]{FFFFE6}-- & \cellcolor[HTML]{FFFFE6}RMSE, RMSE Decrease & \cellcolor[HTML]{FFFFE6}2022 \\ \hhline{~------}  
& \cellcolor[HTML]{FFFFE6}HANet \citep{bi2023hierarchical} & \cellcolor[HTML]{FFFFE6}Multi & \cellcolor[HTML]{FFFFE6}Point & \cellcolor[HTML]{FFFFE6}-- & \cellcolor[HTML]{FFFFE6}MAE, RMSE & \cellcolor[HTML]{FFFFE6}2022 \\ \hhline{~------}  
& \cellcolor[HTML]{FFFFE6}$D^{3}\text{VAE}$ \citep{li2022generative} & \cellcolor[HTML]{FFFFE6}Multi & \cellcolor[HTML]{FFFFE6}Pro & \cellcolor[HTML]{FFFFE6}-- & \cellcolor[HTML]{FFFFE6}MSE, CRPS & \cellcolor[HTML]{FFFFE6}2022 \\ \hhline{~------}  
& \cellcolor[HTML]{FFFFE6}TI-MAE \citep{li2023ti} & \cellcolor[HTML]{FFFFE6}Multi & \cellcolor[HTML]{FFFFE6}Point & \cellcolor[HTML]{FFFFE6}MSE & \cellcolor[HTML]{FFFFE6}MSE, MAE & \cellcolor[HTML]{FFFFE6}2023 \\ \hhline{~------}

\multirow{-14}{*}{\begin{tabular}[c]{@{}c@{}}Encoder\\ -\\ Decoder\end{tabular}}  & \cellcolor[HTML]{B7CDF2}AST \citep{wu2020adversarial} & \cellcolor[HTML]{B7CDF2}Uni & \cellcolor[HTML]{B7CDF2}Pro &\cellcolor[HTML]{B7CDF2} cross-entropy &\cellcolor[HTML]{B7CDF2} Q50, Q90 loss & \cellcolor[HTML]{B7CDF2}2020                  \\ \hline  
                                    & \cellcolor[HTML]{FFCCC9}TFT \citep{lim2021temporal} & \cellcolor[HTML]{FFCCC9}Multi \& Uni &\cellcolor[HTML]{FFCCC9} Prob & \cellcolor[HTML]{FFCCC9}quantile loss &\cellcolor[HTML]{FFCCC9} P50, P90 quantile loss & \cellcolor[HTML]{FFCCC9}2021              \\ \hhline{~------}  
& \cellcolor[HTML]{FFCCC9}Informer \citep{zhou2021informer} & \cellcolor[HTML]{FFCCC9}Multi \& Uni & \cellcolor[HTML]{FFCCC9}Point & \cellcolor[HTML]{FFCCC9}MSELoss & \cellcolor[HTML]{FFCCC9}MSE, MAE & \cellcolor[HTML]{FFCCC9}2021 \\ \hhline{~------}
& \cellcolor[HTML]{FFCCC9}ETSformer \citep{woo2022etsformer} & \cellcolor[HTML]{FFCCC9}Multi \& Uni & \cellcolor[HTML]{FFCCC9}Point & \cellcolor[HTML]{FFCCC9}MSELoss & \cellcolor[HTML]{FFCCC9}MSE, MAE & \cellcolor[HTML]{FFCCC9}2022 \\ \hhline{~------}
& \cellcolor[HTML]{FFCCC9}FEDformer \citep{zhou2022fedformer} & \cellcolor[HTML]{FFCCC9}Multi \& Uni & \cellcolor[HTML]{FFCCC9}Point & \cellcolor[HTML]{FFCCC9}MSELoss & \cellcolor[HTML]{FFCCC9}MSE, MAE, Permutation & \cellcolor[HTML]{FFCCC9}2022 \\ \hhline{~------}
& \cellcolor[HTML]{FFCCC9}TACTiS \citep{drouin2022tactis} & \cellcolor[HTML]{FFCCC9}Multi \& Uni & \cellcolor[HTML]{FFCCC9}Pro & \cellcolor[HTML]{FFCCC9}log-likelihood & \cellcolor[HTML]{FFCCC9}CRPS-Sum, CRPS-means & \cellcolor[HTML]{FFCCC9}2022 \\ \hhline{~------}
& \cellcolor[HTML]{FFCCC9}Autoformer \citep{wu2021autoformer} & \cellcolor[HTML]{FFCCC9}Multi \& Uni & \cellcolor[HTML]{FFCCC9}Point & \cellcolor[HTML]{FFCCC9}L2 loss & \cellcolor[HTML]{FFCCC9}MSE, MAE & \cellcolor[HTML]{FFCCC9}2022 \\ \hhline{~------}
& \cellcolor[HTML]{FFCCC9}NSTformer \citep{liu2022non} & \cellcolor[HTML]{FFCCC9}Multi \& Uni & \cellcolor[HTML]{FFCCC9}Point & \cellcolor[HTML]{FFCCC9}L2 loss & \cellcolor[HTML]{FFCCC9}MSE, MAE & \cellcolor[HTML]{FFCCC9}2023 \\ \hhline{~------}
& \cellcolor[HTML]{FFCCC9}Dateformer \citep{young2022dateformer} & \cellcolor[HTML]{FFCCC9}Multi \& Uni & \cellcolor[HTML]{FFCCC9}Point & \cellcolor[HTML]{FFCCC9}MSE & \cellcolor[HTML]{FFCCC9}MSE, MAE & \cellcolor[HTML]{FFCCC9}2023 \\ \hhline{~------}
  & \cellcolor[HTML]{FFCCC9}Crossformer \citep{zhang2023crossformer} & \cellcolor[HTML]{FFCCC9}Multi \& Uni & \cellcolor[HTML]{FFCCC9}Point & \cellcolor[HTML]{FFCCC9}MSE & \cellcolor[HTML]{FFCCC9}MSE, MAE & \cellcolor[HTML]{FFCCC9}2023 \\ \hhline{~------}
& \cellcolor[HTML]{FFCCC9}Scaleformer \citep{shabani2022scaleformer} & \cellcolor[HTML]{FFCCC9}Multi \& Uni & \cellcolor[HTML]{FFCCC9}Pro & \cellcolor[HTML]{FFCCC9}MSE & \cellcolor[HTML]{FFCCC9}MSE, MAE & \cellcolor[HTML]{FFCCC9}2023 \\ \hhline{~------}
& \cellcolor[HTML]{FFCCC9}BasisFormer \citep{ni2023basisformer} & \cellcolor[HTML]{FFCCC9}Multi \& Uni  & \cellcolor[HTML]{FFCCC9}Point & \cellcolor[HTML]{FFCCC9}MSE & \cellcolor[HTML]{FFCCC9}MSE, MAE & \cellcolor[HTML]{FFCCC9}2023 \\ \hhline{~------}
& \cellcolor[HTML]{FFFFE6}CRT \citep{zhang2022self} & \cellcolor[HTML]{FFFFE6}Multi & \cellcolor[HTML]{FFFFE6}Point & \cellcolor[HTML]{FFFFE6}-- & \cellcolor[HTML]{FFFFE6}ROC-AUC, F1-Score & \cellcolor[HTML]{FFFFE6}2021 \\ \hhline{~------}
& \cellcolor[HTML]{FFFFE6}Pyraformer \citep{liu2021pyraformer} & \cellcolor[HTML]{FFFFE6}Multi & \cellcolor[HTML]{FFFFE6}Point & \cellcolor[HTML]{FFFFE6}MSE & \cellcolor[HTML]{FFFFE6}MSE, MAE & \cellcolor[HTML]{FFFFE6}2022 \\ \hhline{~------}
& \cellcolor[HTML]{FFFFE6}TDformer \citep{zhang2022first} & \cellcolor[HTML]{FFFFE6}Multi & \cellcolor[HTML]{FFFFE6}Point & \cellcolor[HTML]{FFFFE6}MSE & \cellcolor[HTML]{FFFFE6}MSE, MAE & \cellcolor[HTML]{FFFFE6}2022 \\ \hhline{~------}
& \cellcolor[HTML]{FFFFE6}FusFormer \citep{yang2022fusion} & \cellcolor[HTML]{FFFFE6}Multi & \cellcolor[HTML]{FFFFE6}Point & \cellcolor[HTML]{FFFFE6}-- & \cellcolor[HTML]{FFFFE6}RMSE, RMSE Decrease & \cellcolor[HTML]{FFFFE6}2022 \\ \hhline{~------}
& \cellcolor[HTML]{FFFFE6}Scaleformer \citep{shabani2022scaleformer} & \cellcolor[HTML]{FFFFE6}Multi  & \cellcolor[HTML]{FFFFE6}Point & \cellcolor[HTML]{FFFFE6}MSE, Huber, Adaptive loss & \cellcolor[HTML]{FFFFE6}MSE, MAE & \cellcolor[HTML]{FFFFE6}2022 \\ \hhline{~------}
& \cellcolor[HTML]{FFFFE6}Infomaxformer \citep{tang2023infomaxformer} & \cellcolor[HTML]{FFFFE6}Multi & \cellcolor[HTML]{FFFFE6}Pro & \cellcolor[HTML]{FFFFE6}MSELoss & \cellcolor[HTML]{FFFFE6}MSE, MAE & \cellcolor[HTML]{FFFFE6}2023 \\ \hhline{~------}          
                                  
                                    &\cellcolor[HTML]{FFFFE6} PatchTST\citep{nie2022time} &\cellcolor[HTML]{FFFFE6} Multi  & \cellcolor[HTML]{FFFFE6}Point & \cellcolor[HTML]{FFFFE6}Adaptive Loss &\cellcolor[HTML]{FFFFE6} MSE, MAE & \cellcolor[HTML]{FFFFE6}2023               \\ \hhline{~------}  

\multirow{-6}{*}{\centering Transformer}      & \cellcolor[HTML]{FFFFE6}iTransformer\citep{liu2023itransformer} & \cellcolor[HTML]{FFFFE6}Multi  &\cellcolor[HTML]{FFFFE6} Point &\cellcolor[HTML]{FFFFE6} L2 Loss & \cellcolor[HTML]{FFFFE6}MSE, MAE &\cellcolor[HTML]{FFFFE6} 2023                \\ \hhline{~------} 
 
& \cellcolor[HTML]{FFFFE6}MCformer \citep{han2024mcformer} & \cellcolor[HTML]{FFFFE6}Multi  & \cellcolor[HTML]{FFFFE6}Point & \cellcolor[HTML]{FFFFE6}MSE, MAE & \cellcolor[HTML]{FFFFE6}MSE, MAE & \cellcolor[HTML]{FFFFE6}2024 \\ \hhline{~------}
& \cellcolor[HTML]{FFFFE6}SAMformer \citep{ilbert2024unlocking} & \cellcolor[HTML]{FFFFE6}Multi  & \cellcolor[HTML]{FFFFE6}Point & \cellcolor[HTML]{FFFFE6}MSE & \cellcolor[HTML]{FFFFE6}MSE, MAE & \cellcolor[HTML]{FFFFE6}2024 \\ \hhline{~------}
& \cellcolor[HTML]{FFFFE6}TSLANet \citep{eldele2024tslanet} & \cellcolor[HTML]{FFFFE6}Multi  & \cellcolor[HTML]{FFFFE6}Point & \cellcolor[HTML]{FFFFE6}MSE & \cellcolor[HTML]{FFFFE6}MSE, MAE & \cellcolor[HTML]{FFFFE6}2024 \\ \hhline{~------}
& \cellcolor[HTML]{FFFFE6}MASTER \citep{li2024master} & \cellcolor[HTML]{FFFFE6}Multi  & \cellcolor[HTML]{FFFFE6}Point & \cellcolor[HTML]{FFFFE6}MSE & \cellcolor[HTML]{FFFFE6}IC, ICIR, RankIC & \cellcolor[HTML]{FFFFE6}2024 \\ \hhline{~------}
& \cellcolor[HTML]{FFFFE6}TimeSiam \citep{dong2024timesiam} & \cellcolor[HTML]{FFFFE6}Multi  & \cellcolor[HTML]{FFFFE6}Point & \cellcolor[HTML]{FFFFE6}L2, Cross-Entropy & \cellcolor[HTML]{FFFFE6}MSE, MAE, Recall, F1 Score & \cellcolor[HTML]{FFFFE6}2024 \\ \hhline{~------}
& \cellcolor[HTML]{FFFFE6}Chronos \citep{ansari2024chronos} & \cellcolor[HTML]{FFFFE6}Multi  & \cellcolor[HTML]{FFFFE6}Point & \cellcolor[HTML]{FFFFE6}Cross Entropy & \cellcolor[HTML]{FFFFE6}WQL, CRPS, MASE & \cellcolor[HTML]{FFFFE6}2024 \\ \hhline{~------}
& \cellcolor[HTML]{FFFFE6}TimeXer \citep{wang2024timexer} & \cellcolor[HTML]{FFFFE6}Multi  & \cellcolor[HTML]{FFFFE6}Point & \cellcolor[HTML]{FFFFE6}L2 loss & \cellcolor[HTML]{FFFFE6}MSE, MAE & \cellcolor[HTML]{FFFFE6}2024 \\ \hhline{~------}
& \cellcolor[HTML]{FFFFE6}Time-SSM \citep{hu2024time} & \cellcolor[HTML]{FFFFE6}Multi  & \cellcolor[HTML]{FFFFE6}Point & \cellcolor[HTML]{FFFFE6}MSE & \cellcolor[HTML]{FFFFE6}MSE, MAE & \cellcolor[HTML]{FFFFE6}2024 \\ \hhline{~------}
& \cellcolor[HTML]{FFFFE6}SageFormer \citep{zhang2024sageformer} & \cellcolor[HTML]{FFFFE6}Multi  & \cellcolor[HTML]{FFFFE6}Point & \cellcolor[HTML]{FFFFE6}MSE & \cellcolor[HTML]{FFFFE6}MSE, MAE & \cellcolor[HTML]{FFFFE6}2024 \\ \hhline{~------}
& \cellcolor[HTML]{FFFFE6}TIME-LLM \citep{wang2023timemixer} & \cellcolor[HTML]{FFFFE6}Multi  & \cellcolor[HTML]{FFFFE6}Point & \cellcolor[HTML]{FFFFE6}MSE, SMAPE & \cellcolor[HTML]{FFFFE6}MSE, MAE, SMAPE & \cellcolor[HTML]{FFFFE6}2024 \\ \hhline{~------} 
& \cellcolor[HTML]{FFFFE6}CARD \citep{wang2024card} & \cellcolor[HTML]{FFFFE6}Multi  & \cellcolor[HTML]{FFFFE6}Point & \cellcolor[HTML]{FFFFE6}MSE, MAE & \cellcolor[HTML]{FFFFE6}MSE, MAE & \cellcolor[HTML]{FFFFE6}2024 \\ \hhline{~------} 
& \cellcolor[HTML]{B7CDF2}Pathformer \citep{chen2024multi} & \cellcolor[HTML]{B7CDF2}Uni & \cellcolor[HTML]{B7CDF2}Pro & \cellcolor[HTML]{B7CDF2}L1 loss & \cellcolor[HTML]{B7CDF2}MSE, MAE & \cellcolor[HTML]{B7CDF2}2024 \\ \hline
& \cellcolor[HTML]{FFCCC9}ForGAN \citep{koochali2019probabilistic} & \cellcolor[HTML]{FFCCC9}Multi \& Uni & \cellcolor[HTML]{FFCCC9}Pro & \cellcolor[HTML]{FFCCC9}RMSE & \cellcolor[HTML]{FFCCC9}MAE, MAPE, RMSE & \cellcolor[HTML]{FFCCC9}2019 \\ \hhline{~------}  
& \cellcolor[HTML]{FFCCC9}COSCI-GAN \citep{seyfi2022generating} & \cellcolor[HTML]{FFCCC9}Multi \& Uni & \cellcolor[HTML]{FFCCC9}Pro & \cellcolor[HTML]{FFCCC9}Global loss = local + central & \cellcolor[HTML]{FFCCC9}MAE & \cellcolor[HTML]{FFCCC9}2022 \\ \hhline{~------} 
& \cellcolor[HTML]{FFFFE6}RCGAN \citep{esteban2017real} & \cellcolor[HTML]{FFFFE6}Multi & \cellcolor[HTML]{FFFFE6}Pro & \cellcolor[HTML]{FFFFE6}cross-entropy & \cellcolor[HTML]{FFFFE6}AUROC, AUPRC & \cellcolor[HTML]{FFFFE6}2017 \\ \hhline{~------} 
& \cellcolor[HTML]{FFFFE6}TimeGAN \citep{yoon2019time} & \cellcolor[HTML]{FFFFE6}Multi & \cellcolor[HTML]{FFFFE6}Pro & \cellcolor[HTML]{FFFFE6}Unsupervised, Supervised, Reconstruction, Loss & \cellcolor[HTML]{FFFFE6}Discriminative and Predictive Score & \cellcolor[HTML]{FFFFE6}2019 \\ \hhline{~------} 
& \cellcolor[HTML]{FFFFE6}PSA-GAN \citep{2e5681cbd3944bf7a9859994bda18d62} & \cellcolor[HTML]{FFFFE6}Multi & \cellcolor[HTML]{FFFFE6}Point & \cellcolor[HTML]{FFFFE6}Wasserstein loss & \cellcolor[HTML]{FFFFE6}-- & \cellcolor[HTML]{FFFFE6}2022 \\ \hhline{~------} 
& \cellcolor[HTML]{FFFFE6}AEC-GAN \citep{wang2023aec} & \cellcolor[HTML]{FFFFE6}Multi & \cellcolor[HTML]{FFFFE6}Point & \cellcolor[HTML]{FFFFE6}MSE & \cellcolor[HTML]{FFFFE6}ACF, Skew / Kurt, FD & \cellcolor[HTML]{FFFFE6}2023 \\ \hhline{~------}  
& \cellcolor[HTML]{FFFFE6}ITF-GAN \citep{klopries2024itf} & \cellcolor[HTML]{FFFFE6}Multi & \cellcolor[HTML]{FFFFE6}Point & \cellcolor[HTML]{FFFFE6}MSE & \cellcolor[HTML]{FFFFE6}MSE, STS, Pearson, Hellinger, Pred. & \cellcolor[HTML]{FFFFE6}2024 \\ \hhline{~------}  
                                    & \cellcolor[HTML]{FFFFE6}MAGAN \citep{ferchichi2024multi} &\cellcolor[HTML]{FFFFE6} Multi & \cellcolor[HTML]{FFFFE6}Point &\cellcolor[HTML]{FFFFE6}-- & \cellcolor[HTML]{FFFFE6}MAE, MAPE & \cellcolor[HTML]{FFFFE6}2024 \\ \hhline{~------}  
                                    & \cellcolor[HTML]{FFFFE6}TSGAN \citep{xu2024gan} & \cellcolor[HTML]{FFFFE6}Multi & \cellcolor[HTML]{FFFFE6}Point & \cellcolor[HTML]{FFFFE6}-- & \cellcolor[HTML]{FFFFE6}MAE, RMSE, MAPE & \cellcolor[HTML]{FFFFE6}2022 \\ \hhline{~------} 
 
\multirow{-11}{*}{\centering GAN}        & \cellcolor[HTML]{B7CDF2}AST \citep{wu2020adversarial} & \cellcolor[HTML]{B7CDF2}Uni &\cellcolor[HTML]{B7CDF2} Pro & \cellcolor[HTML]{B7CDF2}cross-entropy &\cellcolor[HTML]{B7CDF2} Q50 loss, Q90 loss & \cellcolor[HTML]{B7CDF2}2020                               \\ \hline 
& \cellcolor[HTML]{FFCCC9}ConvLSTM \citep{shi2015convolutional} & \cellcolor[HTML]{FFCCC9}Multi \& Uni & \cellcolor[HTML]{FFCCC9}Point & \cellcolor[HTML]{FFCCC9}cross-entropy & \cellcolor[HTML]{FFCCC9}Rainfall-MSE, CSI, FAR, POD & \cellcolor[HTML]{FFCCC9}2015 \\ \hhline{~------} 
& \cellcolor[HTML]{FFFFE6}Bi-LSTM \citep{du2020multivariate} & \cellcolor[HTML]{FFFFE6}Multi & \cellcolor[HTML]{FFFFE6}Point & \cellcolor[HTML]{FFFFE6}MSE & \cellcolor[HTML]{FFFFE6}MAE, RMSE & \cellcolor[HTML]{FFFFE6}2020 \\ \hhline{~------}  
& \cellcolor[HTML]{FFFFE6}\begin{tabular}[c]{@{}c@{}}\citep{fu2022temporal}\end{tabular} & \cellcolor[HTML]{FFFFE6}Multi & \cellcolor[HTML]{FFFFE6}Point & \cellcolor[HTML]{FFFFE6}MAE & \cellcolor[HTML]{FFFFE6}MAE, RMSE, MAPE & \cellcolor[HTML]{FFFFE6}2022 \\ \hhline{~------} 
& \cellcolor[HTML]{B7CDF2}\citep{asiful2018hybrid} & \cellcolor[HTML]{B7CDF2}Uni & \cellcolor[HTML]{B7CDF2}Point & \cellcolor[HTML]{B7CDF2}L2 loss & \cellcolor[HTML]{B7CDF2}MAE, MSE, MAPE & \cellcolor[HTML]{B7CDF2}2018 \\ \hhline{~------} 
& \cellcolor[HTML]{B7CDF2}TATCN \citep{wang2022tatcn} & \cellcolor[HTML]{B7CDF2}Uni & \cellcolor[HTML]{B7CDF2}Point & \cellcolor[HTML]{B7CDF2}-- & \cellcolor[HTML]{B7CDF2}MAE, RMSE, sMAPE & \cellcolor[HTML]{B7CDF2}2022 \\ \hhline{~------} 

 \multirow{-6}{*}{\centering \begin{tabular}[c]{@{}c@{}}Integrated \\ Module\end{tabular}}                                  & \cellcolor[HTML]{B7CDF2}LST-TCN \citep{sheng2022short}  & \cellcolor[HTML]{B7CDF2} Uni& \cellcolor[HTML]{B7CDF2} Point  &  \cellcolor[HTML]{B7CDF2}Pinball loss  &  \cellcolor[HTML]{B7CDF2}MAPE, RMSE & \cellcolor[HTML]{B7CDF2} 2022              \\\hline
                                    
& \cellcolor[HTML]{FFCCC9}TimesNet \citep{wu2022timesnet} & \cellcolor[HTML]{FFCCC9}Multi \& Uni & \cellcolor[HTML]{FFCCC9}Point & \cellcolor[HTML]{FFCCC9}MSE, SMAPE & \cellcolor[HTML]{FFCCC9}MSE, MAE, SMAPE, MASE & \cellcolor[HTML]{FFCCC9}2022 \\ \hhline{~------} 
& \cellcolor[HTML]{FFCCC9}TreeDRNeT \citep{zhou2022treedrnet} & \cellcolor[HTML]{FFCCC9}Multi \& Uni & \cellcolor[HTML]{FFCCC9}Point & \cellcolor[HTML]{FFCCC9}Lp Regularized Loss & \cellcolor[HTML]{FFCCC9}MSE, MAE & \cellcolor[HTML]{FFCCC9}2022 \\ \hhline{~------}  
& \cellcolor[HTML]{FFCCC9}Triformer \citep{cirstea2022triformer} & \cellcolor[HTML]{FFCCC9}Multi \& Uni & \cellcolor[HTML]{FFCCC9}Point & \cellcolor[HTML]{FFCCC9}-- & \cellcolor[HTML]{FFCCC9}MSE, MAE & \cellcolor[HTML]{FFCCC9}2022 \\ \hhline{~------} 
& \cellcolor[HTML]{FFCCC9}SCINet \citep{liu2022scinet} & \cellcolor[HTML]{FFCCC9}Multi \& Uni & \cellcolor[HTML]{FFCCC9}Point & \cellcolor[HTML]{FFCCC9}L1 loss & \cellcolor[HTML]{FFCCC9}RSE, CORR, MSE, MAE, MAPE, RMSE & \cellcolor[HTML]{FFCCC9}2022 \\ \hhline{~------} 
& \cellcolor[HTML]{FFFFE6}HTSF \citep{duan2023combating} & \cellcolor[HTML]{FFFFE6}Multi & \cellcolor[HTML]{FFFFE6}Pro & \cellcolor[HTML]{FFFFE6}L2 loss, HyperGRU & \cellcolor[HTML]{FFFFE6}MAE, RMSE & \cellcolor[HTML]{FFFFE6}2023 \\ \hhline{~------}  
& \cellcolor[HTML]{FFFFE6}CIPM \citep{yolcu2023novel} & \cellcolor[HTML]{FFFFE6}Multi & \cellcolor[HTML]{FFFFE6}Point & \cellcolor[HTML]{FFFFE6}-- & \cellcolor[HTML]{FFFFE6}RMSE, MAPE, MdRAE & \cellcolor[HTML]{FFFFE6}2023 \\ \hhline{~------}  
 
& \cellcolor[HTML]{FFFFE6}MACN \citep{he2023multi} & \cellcolor[HTML]{FFFFE6}Multi & \cellcolor[HTML]{FFFFE6}Point & \cellcolor[HTML]{FFFFE6}RMSE & \cellcolor[HTML]{FFFFE6}RMSE, MAE & \cellcolor[HTML]{FFFFE6}2023 \\ \hhline{~------} 
& \cellcolor[HTML]{FFFFE6}CasCIFF \citep{zhu2024casciff} & \cellcolor[HTML]{FFFFE6}Multi & \cellcolor[HTML]{FFFFE6}Point & \cellcolor[HTML]{FFFFE6}-- & \cellcolor[HTML]{FFFFE6}MSLE, MAPE & \cellcolor[HTML]{FFFFE6}2024 \\ \hhline{~------} 
& \cellcolor[HTML]{FFFFE6}FCPM \citep{guo2024explainable} & \cellcolor[HTML]{FFFFE6}Multi & \cellcolor[HTML]{FFFFE6}Point & \cellcolor[HTML]{FFFFE6}RMSE & \cellcolor[HTML]{FFFFE6}MAE & \cellcolor[HTML]{FFFFE6}2024 \\ \hhline{~------}                                    

 \multirow{-11}{*}{\centering Cascade}& \cellcolor[HTML]{B7CDF2}N-BEATS \citep{oreshkin2019n}   & \cellcolor[HTML]{B7CDF2}Uni & \cellcolor[HTML]{B7CDF2}Point  & \cellcolor[HTML]{B7CDF2} MAE & \cellcolor[HTML]{B7CDF2}SMAPE, OWA, MASE  &\cellcolor[HTML]{B7CDF2} 2020              \\ \hline
\end{tabular}
\end{table*}

\begin{figure*}[!t]
\centering
\setlength{\abovecaptionskip}{3mm}
\captionsetup{labelsep=none} 
\includegraphics[width=\linewidth,height=\textheight,keepaspectratio]{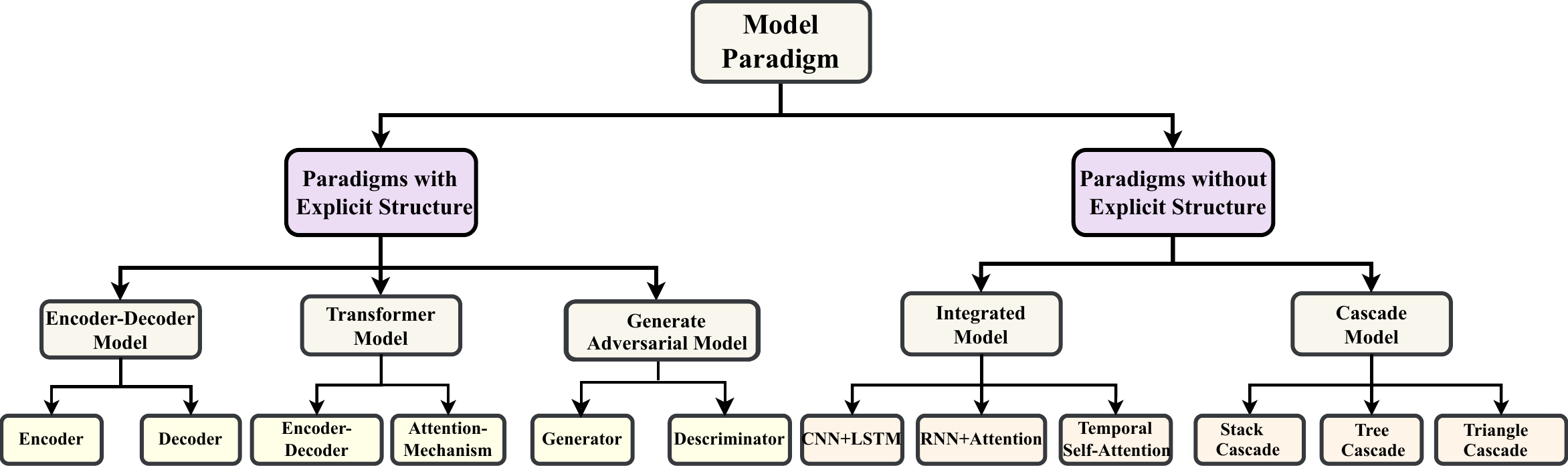}
\caption{ \hspace{1em}The details of five paradigms}
\vspace{-0.1in}
\label{fig2}
\end{figure*}

\section{DTSF Model Architecture}\label{sec3}

Time series data is prevalent in various real-world domains, including energy, transportation, and communication systems. Accurately modeling and predicting time series data plays a crucial role in enhancing the efficiency of these systems. Classical deep learning models (RNN, TCN, Transformer, and GAN) have made significant advancements in TSF  \citep{wu2021autoformer, zhou2022fedformer, woo2022etsformer, zhang2022first}, providing valuable insights for subsequent research.

One of the widely adopted methods is the Recurrent Neural Network (RNN), which utilizes recurrent connections to handle temporal relationships and capture evolving patterns in sequential data. Variants of RNNs, namely Long Short-Term Memory (LSTM) and Gated Recurrent Units (GRU), are specifically designed to address long-term dependencies and effectively capture patterns in long time series. There is a lot of research based on RNNs, DeepAR \citep{salinas2020deepar} leveraged RNN and autoregressive techniques to capture temporal dependencies and patterns in time series data. MQRNN \citep{wen2017multi} exploited the expressiveness and temporal nature of RNNs, the nonparametric nature of Quantile Regression and the efficiency of Direct Multi Horizon Forecasting, proposed a new training scheme named forking-sequences to boost stability and performance. ES-RNN \citep{smyl2020hybrid} proposed a dynamic computational graph neural network with a standard exponential smoothing model and LSTM in a common framework.

In addition to RNNs, Convolutional Neural Networks (CNNs) can also be employed for TSF. By processing time series data as one-dimensional signals, CNNs can extract features from local regions, enabling them to capture local patterns and translational invariance effectively. Notably, Temporal Convolutional Networks (TCNs) represent a prominent example of CNN-based models for time series analysis.

The Temporal Convolutional Network is a classical deep learning model that has garnered widespread attention in time series forecasting due to its ability to effectively capture long-range dependencies. Unlike traditional RNN, TCNs employ convolutional layers with dilated convolutions to expand the receptive field without increasing the number of parameters. This enables TCNs to handle long-range dependencies more efficiently while maintaining computational efficiency \citep{bai2018empirical}. TCNs are particularly useful for time series data with complex temporal patterns, as they can model sequences of varying lengths without suffering from the vanishing gradient problem \citep{deng2019knowledge}. In traffic flow prediction, TCNs have been successfully applied to model the temporal dependencies in sensor data, achieving high accuracy in forecasting traffic conditions \citep{zhao2019deep}. Furthermore, when combined with other techniques such as attention mechanisms and feature extraction layers, TCNs have demonstrated improved performance across various prediction tasks. For instance, integrating TCNs with attention-based models has shown enhanced results in multivariate time series forecasting tasks like electricity load prediction and energy demand forecasting. Overall, TCNs provide a powerful and effective approach to time series forecasting, especially when dealing with long sequences or datasets with complex temporal dependencies.

Another valuable technique is the attention mechanism, which allows models to assign varying weights to different parts of the input sequence. This is particularly beneficial for handling long-term series or focusing on important information at specific time points. Additionally, Generative Adversarial Networks (GANs) can be utilized for TSF. Through adversarial training between a generator and a discriminator, GANs can generate synthetic time series samples and provide more accurate prediction results.

In this section, we dynamically classify existing time series models based on the model architecture dimension. We focus on the internal structural design of the models and categorize the five model architectures into explicit structure paradigms and implicit structure paradigms. Figure \ref{fig2} shows more details of our proposed model classification. Table \ref{DTSF} comprehensively summarizes the models that have made outstanding contributions in recent years. Table \ref{Model_Advantages} selects several key models and provides a detailed analysis of their advantages, disadvantages, application domains, and prediction horizons. The aim is to help readers understand the unique characteristics of each model and guide them in selecting the most suitable model for specific prediction tasks.


\subsection{Model with Explicit Structure}

\subsubsection{Encoder-Decoder Model}
The encoder-decoder model is widely used in the field of deep learning, which appears similar to seq2seq and has an explicit encoder and a decoder. However, seq2seq seems to be described from an application-level perspective, while the encoder-decoder is described at the network level. U-net for medical image segmentation  \citep{ronneberger2015u} and various forms of Transformers are well-known applications.


\begin{table}[ht]
    \renewcommand{\arraystretch}{1.2} 
    \footnotesize
    \captionsetup{labelsep=none} 
    \caption{\hspace{1em} A comparative analysis of time series forecasting models: advantages, disadvantages, applications, and prediction lengths.}
    \vspace{0.03in}
    \label{Model_Advantages}
    \begin{tabularx}{\columnwidth}{
        >{\centering\arraybackslash}m{2cm}|
        >{\centering\arraybackslash}m{3cm}
        >{\centering\arraybackslash}m{3cm}
        >{\centering\arraybackslash}m{1.5cm}
        >{\centering\arraybackslash}m{1.5cm}
        >{\raggedright\arraybackslash}p{5cm}} 
        
\hline
Model      & Advantages                                                                                     & Disadvantages                                & Applications                      & Prediction Horizon \\ \hline
Informer \citep{zhou2021informer}   & \begin{tabular}[c]{@{}l@{}}Effcient; Strong repre-\\-sentation; Good \\generalization\end{tabular} & Sensitive to data shifts                     & Energy; Weather & Long               \\ \hline
HANet \citep{bi2023hierarchical}      & Capture complex dependencies; Flexible for multivariate data                                   & High complexity                              &  Weather; Ecology & Long               \\ \hline
Autoformer  \citep{wu2021autoformer} & Efficient; Good information                                                                     & High complexity; Depend on data periodicity  & Finance; Energy; Electricity; Traffic; Weather; Healthcare & Long               \\ \hline
ETSformer \citep{woo2022etsformer} & Combine traditional methods with Transformer; Adaptive time window                             & High computational cost; Requires large data & Finance; Energy; Electricity; Traffic; Weather; Healthcare & Short              \\ \hline
FEDformer \citep{zhou2022fedformer}  & Frequency enhancement; Better flexibility for long-term forecasts                             & High complexity; Large data needed           & Finance; Energy; Electricity; Traffic; Weather; Healthcare & Long               \\ \hline
TreeDRNet \citep{zhou2022treedrnet} & Capture time dynamics; Efficient training with joint networks                                & High complexity; Need large data.           & Finance; Energy; Electricity; Traffic; Weather; Healthcare & Long               \\ \hline
TATCN \citep{wang2022tatcn}    & Capture temporal dependencies; Extract local patterns.                                       & High computational cost; Data dependence.    & Electricity; Healthcare & Short              \\ \hline
 \end{tabularx}
\end{table}

In this context, the classic Seq2Seq model stands as one of the most representative Encoder-Decoder architectures. It uses Long Short-Term Memory networks as both the encoder and decoder to map input sequences to output sequences, making it particularly suitable for multi-step forecasting tasks \citep{sutskever2014sequence}. Additionally, LSTM and GRU are classic models for time series data modeling, capable of capturing long-term dependencies, and have demonstrated excellent performance in various time series forecasting tasks, such as financial forecasting and weather prediction \citep{cho2014learning}. In contrast to traditional RNNs, TCN leverage convolutional layers to address long-term dependency issues, achieving strong results in several time series forecasting applications, particularly in traffic flow prediction and weather forecasting \citep{bai2018empirical}. Moreover, the Bi-directional Encoder-Decoder model, which utilizes bidirectional LSTM, captures both past and future time information, further enhancing the model's forecasting accuracy \citep{cheng2022dual}. These classic Encoder-Decoder models, with their ability to automatically learn complex patterns in time series data, have become essential tools in time series forecasting tasks.

Encoder-decoder has also been extensively and successfully applied in the field of TSF. For instance,  \citet{perslev2019u} was inspired by U-net  \citep{ronneberger2015u} and designed a time fully convolutional network called U-Time based on the U-net architecture. U-Time maps arbitrarily long sequential inputs to label sequences on a freely chosen time scale. The overall network exhibits a U-shaped architecture with highly symmetric encoder and decoder components. We believe that the high degree of symmetry in the architecture is because the proposed network's input and output exist in the same space. The encoder maps the input into another space, and the decoder should map back from this space. Therefore, the network architecture is theoretically highly symmetric.

There are many highly symmetric encoder-decoder network architectures, as well as cases where the encoder and decoder are asymmetric. The most typical example is the Transformer architecture  \citep{zhou2021informer, wu2021autoformer, zhou2022fedformer, yang2022fusion}. It can be observed that the decoder differs from the encoder and receives input. This encoder-decoder architecture is considered to require additional information for assistance to perform better.

Likewise, \cite{guo2023dynamic} proposed an asymmetric encoder-decoder learning framework where the spatial relationships and time-series features between multiple buildings are extracted by a convolutional neural network and a gated recurrent neural network to form new input data in the encoder. The decoder then makes predictions based on the input data with an attention mechanism.

There are some other examples of encoder-decoder here as well. In  \citet{bi2023hierarchical}, a novel hierarchical attention network (HANet) for the long-term prediction of multivariate time series was proposed, which also includes an encoder and a decoder. However, the encoder and decoder architectures are noticeably different. That is to say, the encoder and decoder are asymmetric. There are also network architectures that explicitly involve an encoder but lack an explicit decoder\citep{eldele2021time}.

\begin{figure}[!t]
\centering
\setlength{\abovecaptionskip}{3mm}
\captionsetup{labelsep=none} 
\includegraphics[width=\linewidth]{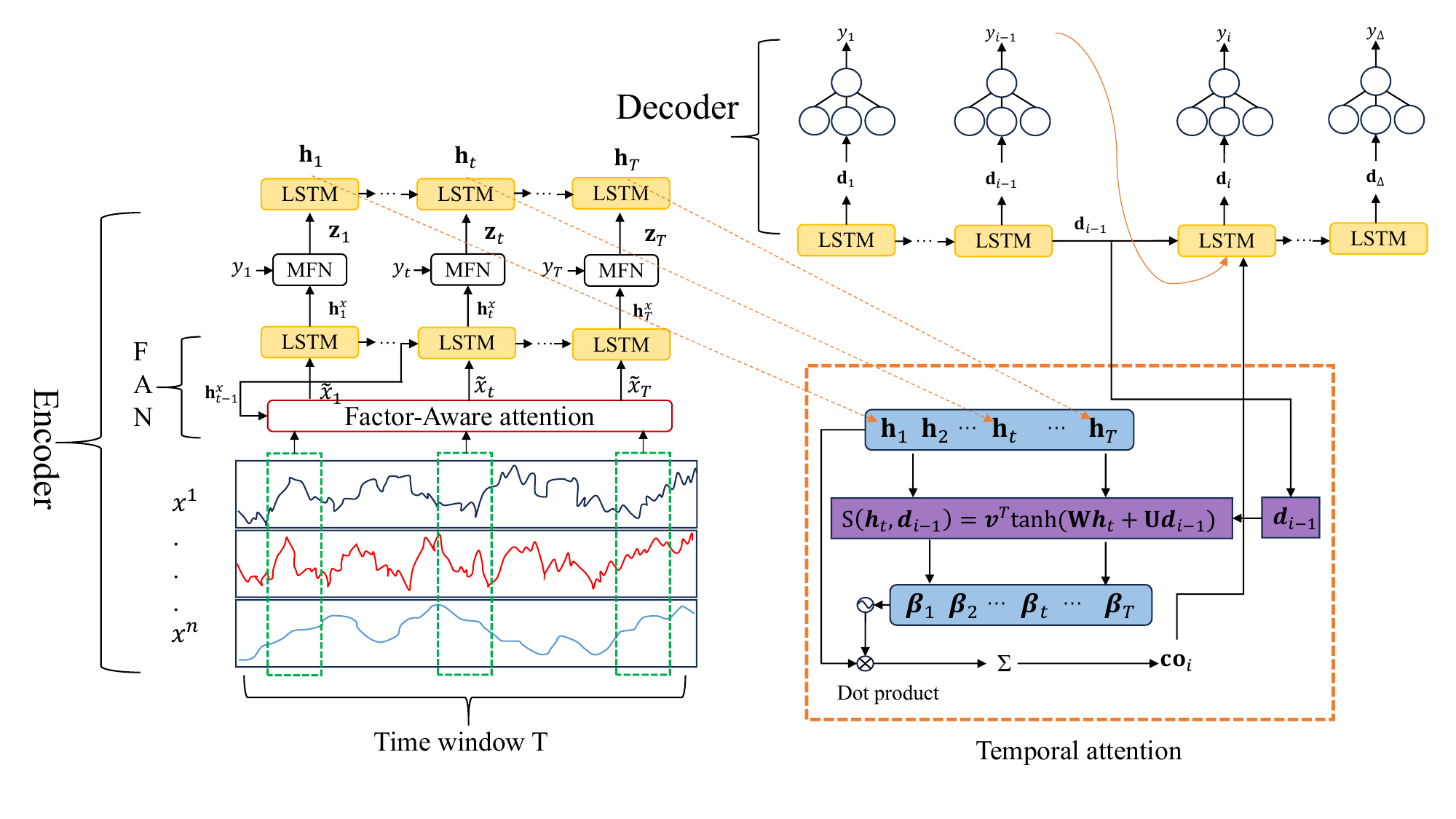}
\caption{\hspace{1em} The overview of HANet model}
\vspace{-0.1in}
\label{fig3}
\end{figure}
\subsubsection{Transformer Model}
With the remarkable performance of Transformer in computer vision and Natural Language Processing (NLP) domains, they have also been applied to the field of TSF and have shown great promise. The main architecture of the Transformer includes the attention mechanism and the encoder-decoder architecture.

However, applying Transformer to TSF tasks is not without challenges and limitations. Recent studies have highlighted several issues, such as the inability to directly handle Long Sequence Time Forecasting (LSTF), including quadratic time complexity, high memory usage, and inherent limitations of the encoder-decoder architecture. To address these limitations, Informer \citep{zhou2021informer} was an efficient Transformer-based architecture specifically designed for LSTF. This architecture utilizes the ProbSparse self-attention mechanism, which reduces the time complexity and memory usage to O(LlogL). From the network architecture perspective, it is evident that Informer's  architecture\citep{zhou2021informer} closely resembles the vanilla Transformer, consisting of an encoder and a decoder. The encoder receives the input, and the decoder receives the output from the encoder as well as the input, with the addition of zero-padding in the parts to be predicted. The self-attention mechanism is replaced with the ProbSparse self-attention mechanism. TFT \citep{lim2021temporal} proposed other architectural improvements to improve accuracy and computational complexity, which integrates high-performance multi-horizon forecasting with interpretable insights into temporal dynamics, capturing temporal relationships at different scales by employing recurrent layers for local processing and interpretable self-attention layers for long-term dependencies. 
\begin{figure}[ht]
\centering
\setlength{\abovecaptionskip}{3mm}
\captionsetup{labelsep=none} 
\includegraphics[width=\linewidth]{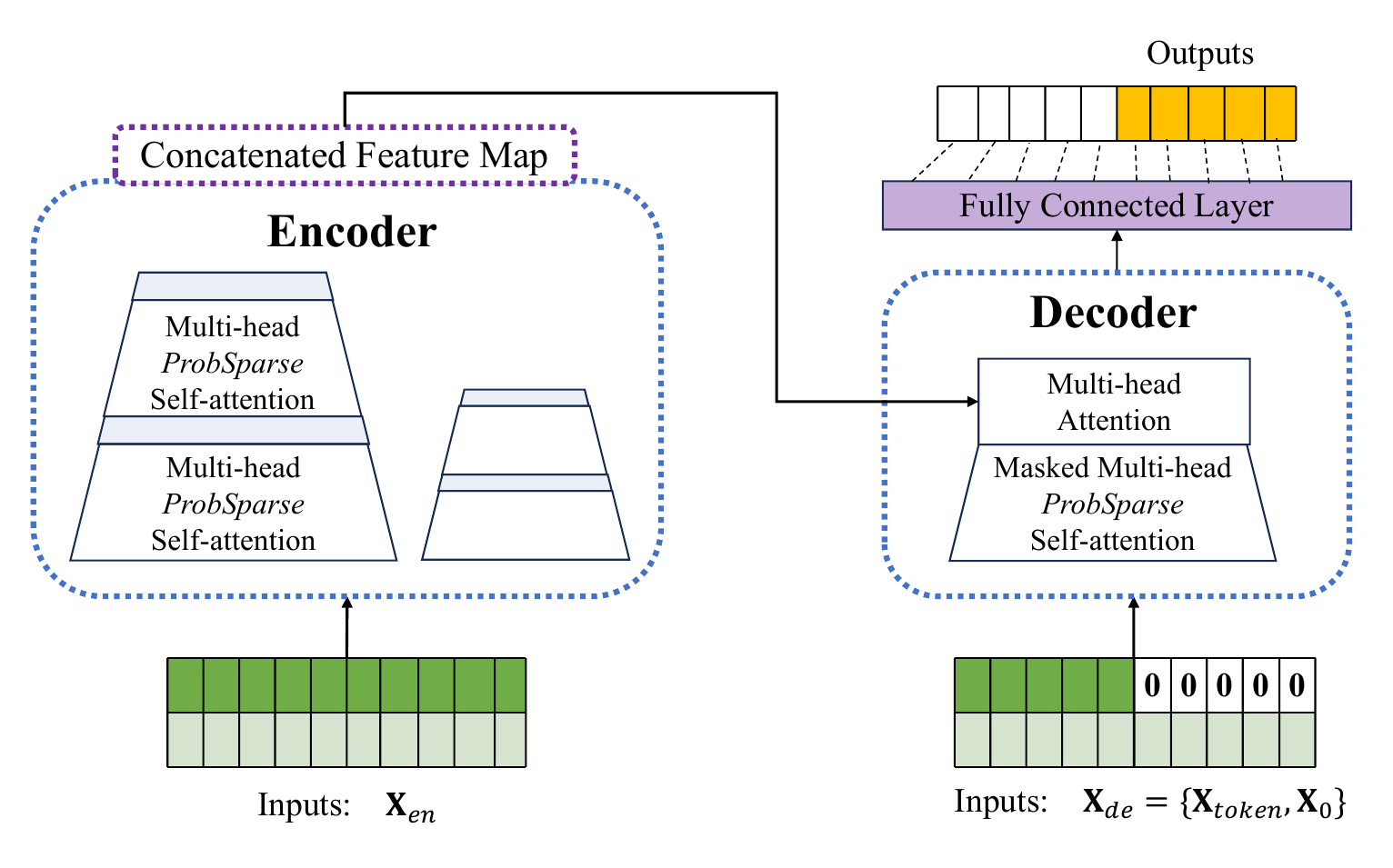}
\caption{\hspace{1em}The overview of Informer model}
\vspace{-0.1in}
\label{fig4}
\end{figure}

Autoformer \citep{wu2021autoformer}, on the other hand, argues that previous Transformer-based prediction models (e.g., Informer \citep{zhou2021informer}) mainly focused on improving self-attention for sparse versions. While significant performance improvements were achieved, they sacrificed the utilization of information. One of the reasons why Transformer cannot be directly applied to LSTF is the complex characteristics of time series data. Without special design, traditional attention mechanisms struggle to model and learn these characteristics. Autoformer \citep{wu2021autoformer} adopts decomposition as a standard approach for time series analysis \citep{makridakis1978time, cleveland1990stl}, as it is believed that decomposition can untangle the intertwined time patterns and highlight the intrinsic properties of time series. Autoformer \citep{wu2021autoformer} introduces a novel decomposition architecture with autocorrelation mechanisms, which is different from the conventional series decomposition preprocessing. In terms of the network architecture, it follows a macro architecture similar to Transformers, Informer, and other architectures. The difference lies in the input to the Decoder, which is no longer the original input but rather sub-sequences obtained through time series decomposition, including seasonal and trend dimensions.

\begin{figure}[!t]
\centering
\setlength{\abovecaptionskip}{3mm}
\captionsetup{labelsep=none} 
\includegraphics[width=\linewidth]{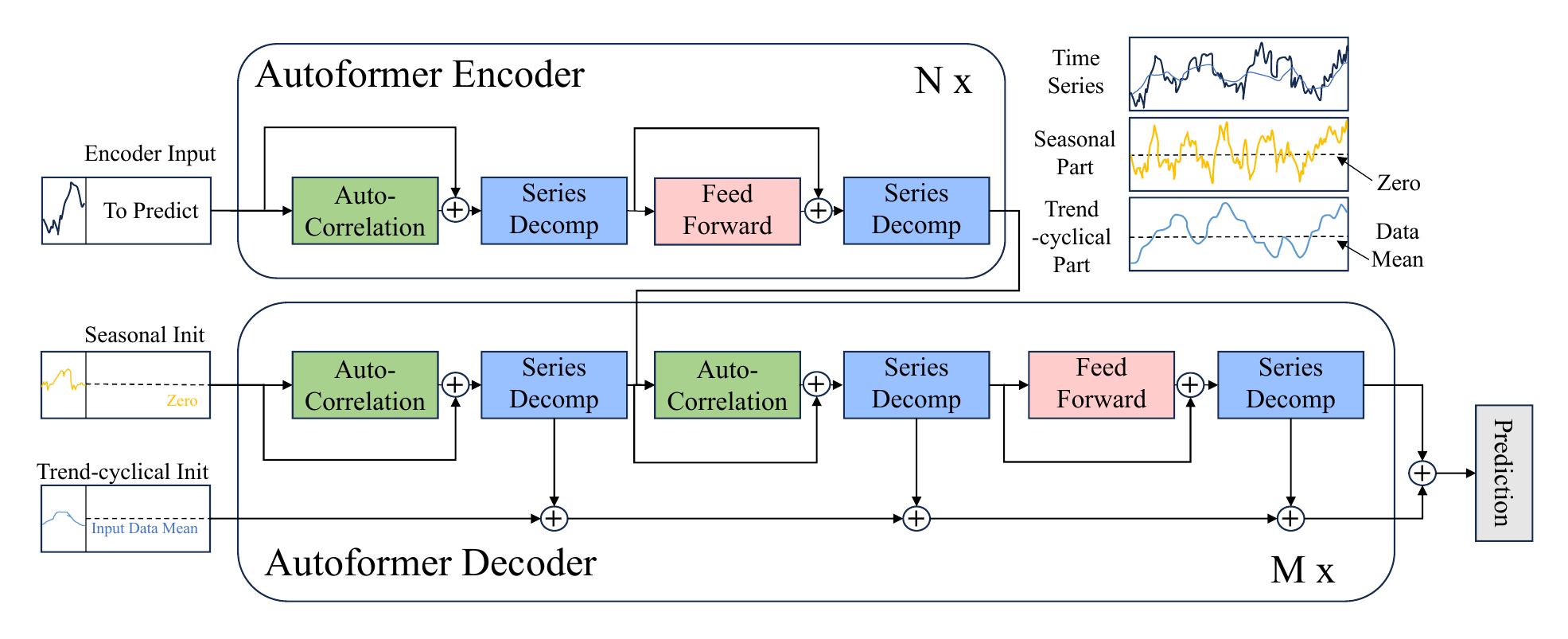}
\caption{ \hspace{1em}The overview of Autoformer model}
\vspace{-0.1in}
\label{fig5}
\end{figure}

In time series forecasting tasks, many researchers prefer to divide long time series into smaller segments to help Transformer models focus more effectively on local temporal features.  This approach enhances the model’s ability to learn local patterns while reducing computational burden. TSMixer \citep{ekambaram2023tsmixer} adopts a similar strategy by partitioning time series data into multiple patches and then processing these patches through MLP-based layers to extract features.  This approach, akin to patch-based methods in computer vision, enables the model to capture local features effectively while reducing computational complexity and memory requirements in time series forecasting tasks. \cite{zhang2023multi} proposed a novel Transformer-based multivariate time series modeling approach in their work, MTPNet. It achieves modeling of temporal information at arbitrary granularities by simultaneously embedding temporal and spatial dimensions of the Seasonal part of the time series decomposition patches.

There are further works addressing Transformers in the context of TSF. ETSformer \citep{woo2022etsformer} argues that the sequence decomposition used by Autoformer makes simplified assumptions and is insufficient to properly model complex trend patterns. Considering that seasonal patterns are more easily identifiable and detectable, ETSformer designs exponential smoothing attention (ESA) and frequency attention (FA) mechanisms. The network architecture decomposes the time series into interpretable sequence components such as level, growth, and seasonality. FEDformer combines Transformers with seasonal-trend decomposition methods. The decomposition method captures the global profile of the time series, while the Transformer captures more detailed architectures, making it a frequency-enhanced Transformer.

These studies demonstrate the ongoing efforts in leveraging Transformers for TSF and the development of specialized architectures and mechanisms to overcome the challenges and limitations associated with applying Transformers to this domain.\par

\subsubsection{Generative Adversarial Model}
GAN (Generative Adversarial Networks) has attracted significant attention since its introduction as a generative model consisting of an explicit structure including a discriminator and a generator. While GANs have been widely used in the field of computer vision, their application in TSF has been relatively limited. The reason for this limited usage is speculated to be the availability of alternative metrics such as CRPS (Continuous Ranked Probability Score) that can measure the quality of generated samples  \citep{benidis2022deep}.

In the existing literature on GAN-based TSF, most studies focus on generating synthetic time series datasets  \citep{yoon2019time, esteban2017real, takahashi2019modeling}. The discriminator is trained to distinguish between real and generated time series data, with the goal of producing synthetic data that is indistinguishable from real data. TimeGAN \citep{yoon2019time}, a GAN-based network architecture, was proposed to generate realistic time series data by leveraging the flexibility of unsupervised models and the control of supervised models. It utilizes an embedding function and a recovery function to extract high-dimensional features from time series data, which are then fed into the sequence generator and sequence discriminator for adversarial training. Another study proposed a GAN-based network architecture using Recurrent Neural Networks (RNNs) to generate real-valued multidimensional time series \citep{takahashi2019modeling}. The study introduced two variations, Recursive GAN (RGAN) and Recursive Conditional GAN (RCGAN), where RGAN generates real-valued data sequences, and RCGAN generates sequences conditioned on specific inputs. The discriminators and generators of both RGAN and RCGAN are based on simple RNN architectures.

Furthermore, a deep neural network-based approach was proposed for modeling financial time series data \citep{takahashi2019modeling}. This approach learns the properties of the data and generates realistic data in a data-driven manner, while preserving statistical characteristics of financial time series such as nonlinear predictability, heavy-tailed return distributions, volatility clustering, leverage effect, coarse-to-fine volatility correlations, and asymmetric return/loss patterns.

\begin{figure}[ht]
\centering
\setlength{\abovecaptionskip}{3mm}
\captionsetup{labelsep=none} 
\includegraphics[width=\linewidth]{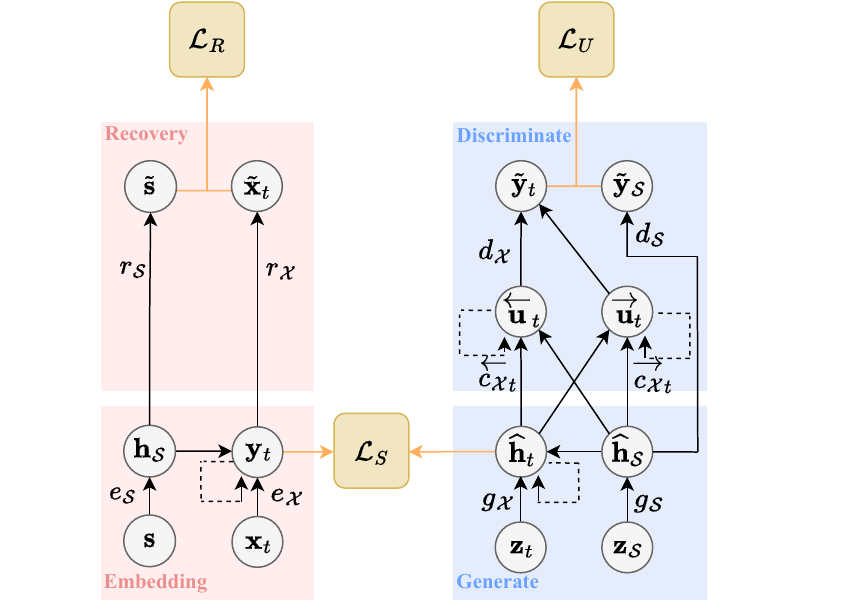}
\caption{\hspace{1em}The overview of TimeGAN model}
\vspace{-0.1in}
\label{fig6}
\end{figure}

These studies highlight the application of GANs in TSF, specifically in generating synthetic time series data and capturing the characteristics of real-world time series data.

\subsection{Model without Explicit Structure}

\subsubsection{\bf {Integrated Model}}
As widely known, recurrent neural networks (RNNs) are often considered suitable for sequence modeling, and the chapter on sequence modeling in classic deep learning textbooks is titled ``Sequence Modeling: Recurrent and Recursive Nets'' \citep{heaton2018ian}. Time series naturally falls within the realm of sequence modeling tasks, and therefore, RNNs, LSTM, GRU, and similar models are expected to be applicable to solve time series-related tasks. However, convolutional architectures have achieved state-of-the-art accuracy in tasks such as audio synthesis, word-level language modeling, and machine translation  \citep{bai2018empirical}, which has garnered significant attention and led to inquiries on how to apply convolutional architectures in the domain of sequences. Integrated models have emerged as a solution.
\begin{figure}[ht]
\centering
\setlength{\abovecaptionskip}{3mm}
\captionsetup{labelsep=none} 
\includegraphics[width=\linewidth]{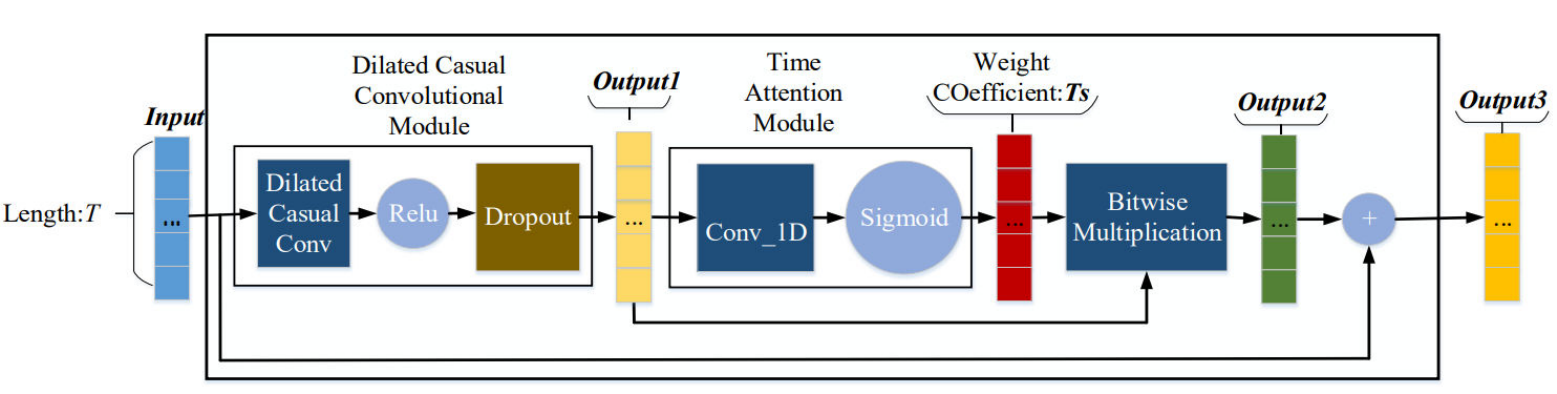}
\caption{\hspace{1em} The overview of TATCN model}
\vspace{-0.1in}
\label{fig7}
\end{figure}

Integrated models can combine the strengths of individual model architectures, with each focusing on learning features it excels at, resulting in improved performance. For example, convolutional architectures excel at learning local feature patterns, while recurrent architectures excel at learning temporal dependencies between nodes. Integrated models have also found various applications in time series tasks  \citep{bai2018empirical, shi2015convolutional, asiful2018hybrid}. In  \citep{shi2015convolutional}, precipitation forecasting was modeled as a spatio-temporal sequence prediction problem, where a convolutional architecture was designed to replace fully connected layers in LSTM for sequence modeling, effectively leveraging the advantages of both convolutional and recurrent architectures. Similarly,  \cite{asiful2018hybrid} integrated multiple network architectures, namely LSTM and GRU, for stock prediction. In this model, the input was first fed into the LSTM layer, then into the GRU layer, and finally into a dense network.\par

\subsubsection{Cascade Model}
Cascade networks, which are widely used in deep neural networks, especially in Computer Vision (CV) domain  \citep{cai2018cascade}, have multiple applications. A cascade network typically consists of multiple components, each serving a different function, collectively forming a deeper and more powerful network model. The components in a cascade model can be either identical or different. When the components are different, each component has a specific role and function. If the components are the same, it means that a particular module or the entire network is repeated several times. When the same component is repeated multiple times, its concept is somewhat similar to the iterative approach used in solving optimization problems.

\begin{figure}[t]
\centering
\setlength{\abovecaptionskip}{3mm}
\captionsetup{labelsep=none} 
\includegraphics[width=\linewidth]{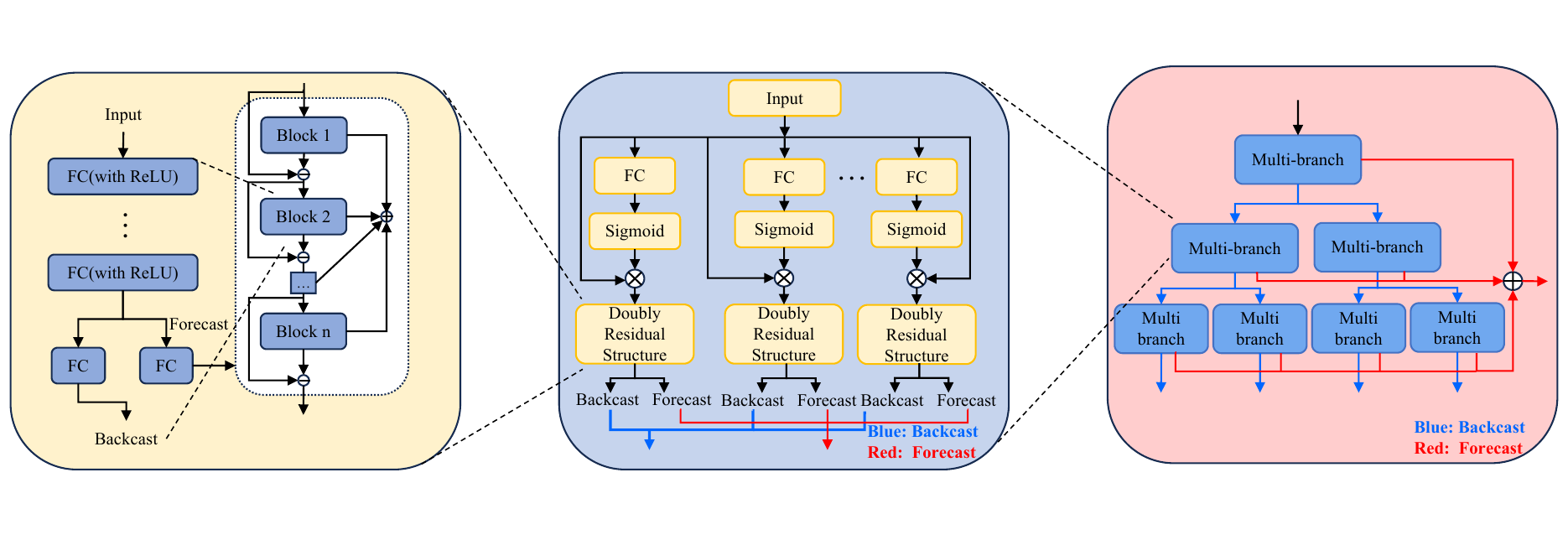}
\caption{\hspace{1em} The overview of TreeDRNet model}
\vspace{-0.1in}
\label{fig8}
\end{figure}

In the field of TSF, there are not many works specifically known for their cascade models. However, the concept of cascade is widely applied in various network model architectures. Firstly, stacking multiple identical modules or the entire network can be considered as utilizing the cascade idea, as seen in the Transformer series  \citep{makridakis1978time, cleveland1990stl,li2019enhancing, de2011forecasting}. Additionally, some models \citep{zhou2022treedrnet} incorporate specially designed cascade approaches to ensure the flow of information in a specific manner, thereby achieving unique effects.\par

\section{Series Components and Enhanced Feature Extraction Methodology}\label{sec4}

In the previous sections, we have provided a comprehensive overview of five prominent paradigms for constructing DTSF models. These paradigms offer researchers a concise pathway to understanding and building DL models. However, a macroscopic understanding and construction of DTSF models alone is insufficient. This chapter delves into the methodological aspects of learning temporal features, which enable models to better capture the underlying representations of the data, emphasizing a pre-training, decomposition, extraction, and refinement process that aligns closely with the intrinsic nature of data.

The chapter is divided into two parts. It begins by dissecting the constituents of time series data in the real world. Subsequently, it proceeds to provide an in-depth exploration of four well-established feature extraction methods with strong theoretical foundations and notable performance in the field. These methods facilitate a richer understanding of time series data and its essential features.\par

\subsection{Components of a Time Series}
\begin{figure}[!t]
\centering
\setlength{\abovecaptionskip}{3mm}
\captionsetup{labelsep=none} 
\includegraphics[width=\linewidth]{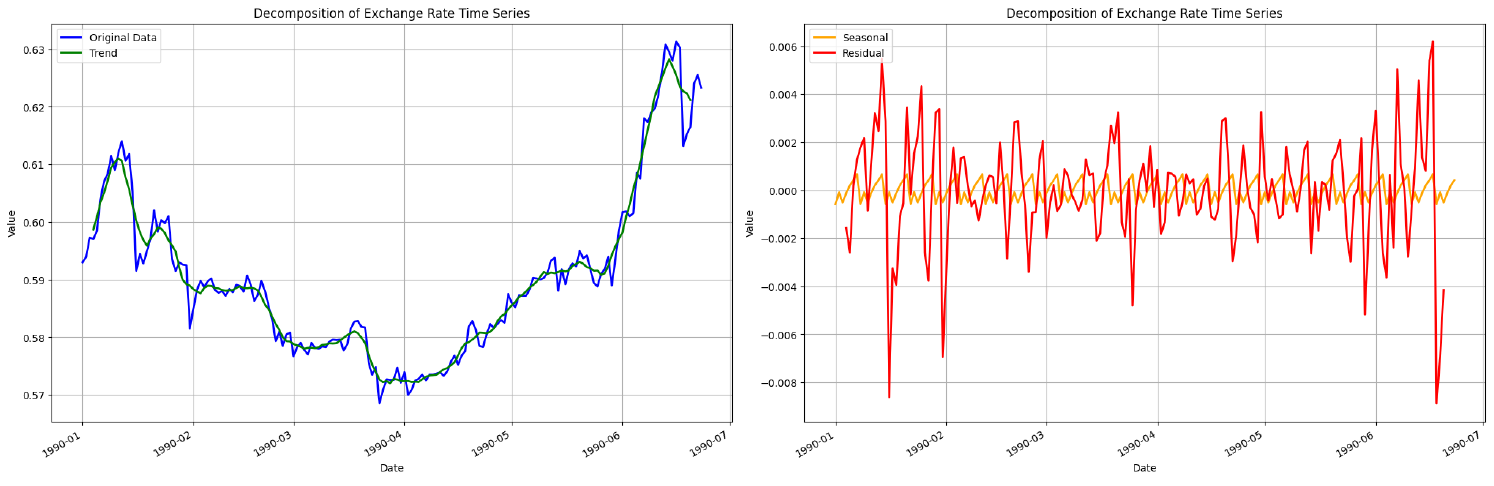}
\caption{ \hspace{1em}Components of the time series. The data is sourced from the Exchange-Rate dataset spanning from January 1, 1990, to June 23, 1990. The blue line represents the original data, the green indicates the trend, the yellow represents seasonality, and the red signifies the residuals}
 \vspace{-0.1in}
\vspace{-0.1in}
\label{fig9}
\end{figure}

In general, time series data can be decomposed into three main components: trend, seasonality, and residuals or white noise \citep{shumway2000time}, as illustrated in Figure \ref{fig9}.

\subsubsection{Trend} Represents the long-term changes in the time series data and reflects the overall growth or decline of the data over an extended period \citep{montgomery2015introduction}. For example, the increase in population over the years exhibits an upward trend \citep{adhikari2013introductory}, and the growing wind power generation during multiple windy seasons can also be considered an upward trend.

\subsubsection{Seasonality} Refers to the periodic variations observed in time series data, often caused by seasonal, monthly, weekly, or other time unit influences. For instance, the number of tourists and ice cream sales tend to increase during long vacations or in the summer.

\subsubsection{Residuals} Represent the part of the data that cannot be explained by the trend and seasonality components \citep{maronna2019robust}. They capture the random fluctuations or noise remaining after the decomposition of trend and seasonality. Residuals reflect the short-term fluctuations and irregularities that have not been modeled in the time series data. Additionally, residuals exhibit some autocorrelation, which can help us identify and adjust for potential flaws in the model, further enhancing the quality and reliability of forecasting.

In the real world, time series data contains discrete information and is non-stationary, meaning that its mean and variance are not constant over time. By decomposing the data into its constituent parts, we gain a better understanding of the data's structure, identify long-term trends and periodic variations, and distinguish them from random noise. These decomposition components aid in making more accurate forecasts, uncovering hidden patterns, extracting useful information, and providing insights into the mechanisms and regularities underlying the time series data.\par

\subsection{Methodology for Enhanced Feature Extraction}

Numerous studies have been dedicated to improving the model architecture and refining its components in DTSF. These studies aim to enhance the predictive performance of models by optimizing or replacing the methods used for extraction and feature learning. To achieve accurate predictions, it is crucial to learn time series representation features thoroughly, and sufficient information is essential for training high-quality model parameters.

In recent years, influential works on DTSF have shown significant changes in data processing and component modeling. Notably, decomposing time series into its major components for analysis has been a primary focus, facilitating a more comprehensive exploration of trends and seasonal dimensions. Furthermore, transforming time-domain data into the frequency domain has proven to be more effective in feature differentiation. Additionally, exploring non-end-to-end approaches and devising suitable data pre-training methods to address the potential mismatch between the target task and the data is also a valuable consideration. In the following sections, we will introduce the primary methodologies for enhancing feature extraction and learning in DTSF.\par
\subsubsection{Dimension Decomposition}
Dimension decomposition plays a vital role in the realm of TSF. It involves breaking down the data into its constituent dimensions or components, such as trends, seasonal patterns, and residuals.

\begin{figure}[!t]
\centering
\setlength{\abovecaptionskip}{3mm}
\captionsetup{labelsep=none} 
\includegraphics[width=\linewidth]{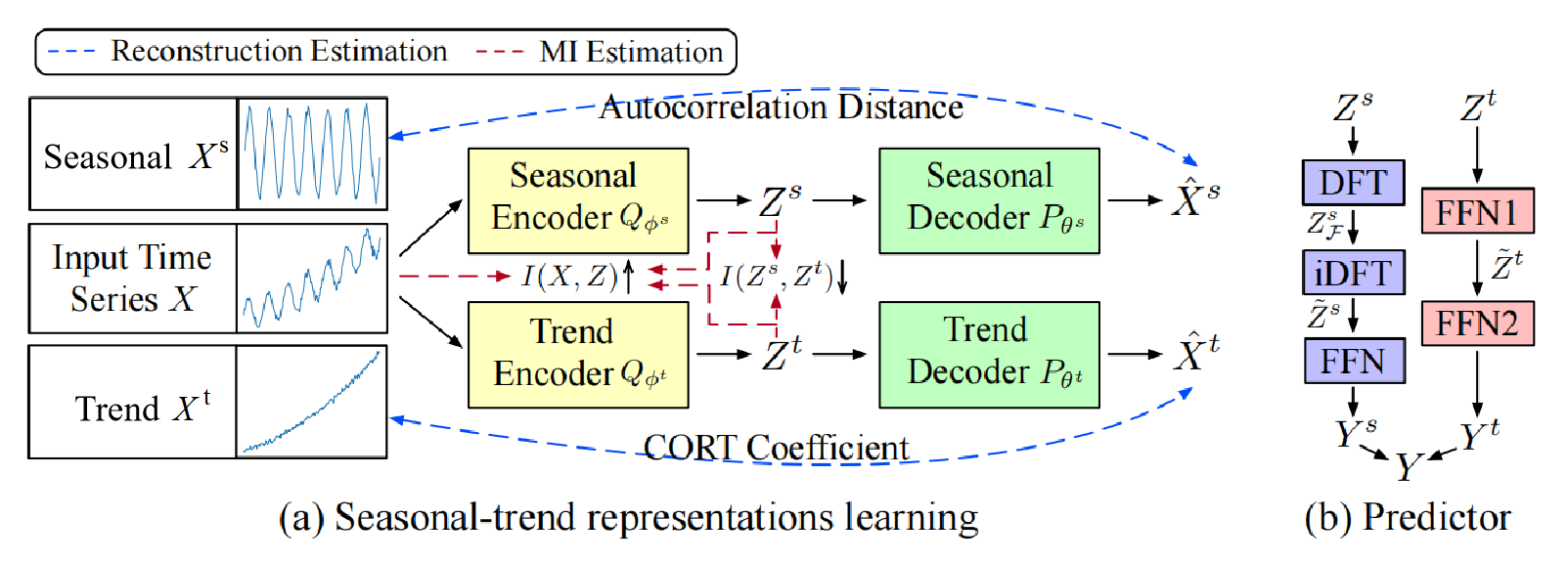}
\caption{\hspace{1em} The overview of LaST model}
\vspace{-0.1in}
\label{fig10}
\end{figure}

In current research, some works have integrated encoder-decoder architectures with seasonal-trend decomposition \citep{wu2021autoformer,zhou2022fedformer,zhang2022first,wang2022learning, zhu2023drcnn, tang2023infomaxformer, cao2023inparformer, peng2023metaformer}. \cite{wu2021autoformer} in the similar work, devised an internal decomposition block to endow deep forecasting model with intrinsic progressive decomposition capability. Subsequently, \cite{zhou2022fedformer} proposed a seasonal-trend-based frequency enhanced decomposition Transformer architecture in the FEDformer framework. Additionally,  \cite{wang2022learning} introduced the LaTS model, leveraging variational inference to unravel latent space seasonal trend features, and \cite{zhang2022first} presented the TDformer model, using MLP to model trends and Fourier attention to simulating seasonality. Notably, \cite{zhu2023drcnn} designed an approach to decompose input sequences into trend and residual components across multiple scales, which summed the learned features as the model output. In recent work, the challenge of capturing outer-window variations was overcome by employing contrastive learning and an enhanced decomposition architecture \citep{park2024self}. It is observed that decomposition networks can significantly benefit contrastive loss learning of long-term representations, thereby enhancing the performance of long-term forecasting.

The significance of dimension decomposition lies in its ability to delve into and capture the inherent components or dimensions within time series data. On one hand, it aids in isolating and extracting latent patterns in time series data for identification and analysis. On the other hand, it isolates individual features that influence the overall behavior, allowing for a more focused analysis of each constituent part. This contributes to understanding the impact of each feature on the overall time series. Furthermore, decomposing data dimensions enhances the interpretability of TSF models, which facilitates a better understanding of the influence of different components on overall temporal behavior. As a relatively universal method in time series analysis, dimension decomposition plays a foundational yet crucial role in enhancing feature extraction methodologies.\par

\subsubsection{Time-Frequency Conversion}
The time-frequency domain conversion plays a crucial role in deep learning-based time series forecasting tasks. It refers to converting the time-domain data into its frequency-domain representation, enabling a more effective analysis of the frequency, spectral characteristics, and dynamic variations within time series data.

\begin{figure}[!t]
\centering
\setlength{\abovecaptionskip}{3mm}
\captionsetup{labelsep=none} 
\includegraphics[width=\linewidth]{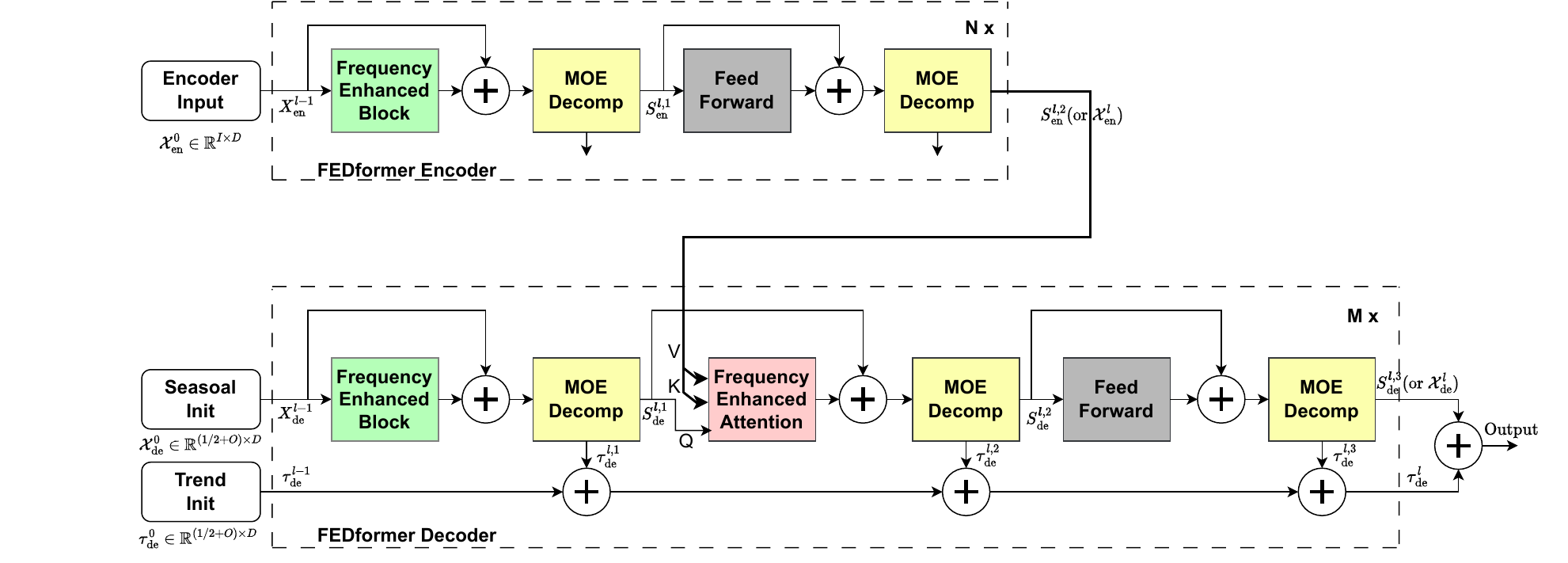}
\caption{\hspace{1em} The overview of FEDformer model}
\vspace{-0.1in}
\label{fig11}
\end{figure}

In current research, the time-frequency domain conversion finds extensive application in the preprocessing and feature extraction of time series data \citep{kourentzes2014improving, chen2023joint, sun2022fredo}. This method reveals the components of the data at different frequencies and aids in identifying repetitive patterns, periodic trends, and frequency-domain features such as seasonal patterns or periodic oscillations \citep{zhou2022fedformer}. Converting time series data into spectrograms provides an overview of the data's distribution in the frequency domain, facilitating the identification of major frequency components and the shape of the spectrum. This is particularly valuable for capturing the overall spectral characteristics of signals and the primary fluctuation patterns across frequencies. In their work, \citep{cao2020spectral} employ StemGNN to jointly capture inter-sequence correlations and temporal dependencies in the spectral domain for multivariate time series forecasting. In recent work, \cite{yi2023frequency} proposed a simple yet effective time series forecasting architecture, named FreTS, based on Frequency-Domain MLP. It primarily consists of two stages, domain conversion and frequency learning, which enhance the learning of channel and temporal correlations across both inter-series and intra-series scales.

Furthermore, employing time-frequency domain conversion can help reduce the impact of noise and interference \citep{zhou2022film,gu2021efficiently}. In specific time series forecasting scenarios, noise may affect the data, resulting in a decline in the model's predictive performance. In the FiLM model, \cite{zhou2022film} introduced a Frequency Enhancement Layer to address this issue. They achieved noise reduction by combining Fourier analysis and low-rank matrix approximation, which minimized the influence of noise signals and mitigated overfitting problems. Apparently, converting time-domain data into the frequency-domain, along with operations like filtering and denoising in the frequency domain, proves effective in lessening the impact of noise.

The importance of time-frequency domain conversion lies in providing a comprehensive and detailed approach to data analysis, which is capable of unveiling the hidden frequency characteristics and dynamic changes within time series. This technique has been widely employed in the domain of TSF, representing a crucial methodology for enhancing predictive performance and comprehending the intricacies of time series data.

\subsubsection{Pre-training}
Compared to natural language, temporal data exhibits lower information density, necessitating longer sequences to capture temporal patterns. Additionally, temporal data also exist challenges such as temporal dynamics, rapid evolution, and the presence of both long and short-term effects. Due to potential mismatches between pre-training and target domains, downstream performance might suffer. Recent endeavors in TSF involve novel attempts at self-supervised and unsupervised pre-training, yielding promising results \citep{rebjock2021online, sun2021adjusting, sarkar2020self, cheng2020subject}. In certain scenarios, the adoption of sampling pre-training methods could be considered.

\begin{figure}[ht]
\centering
\setlength{\abovecaptionskip}{3mm}
\captionsetup{labelsep=none} 
\includegraphics[width=\linewidth]{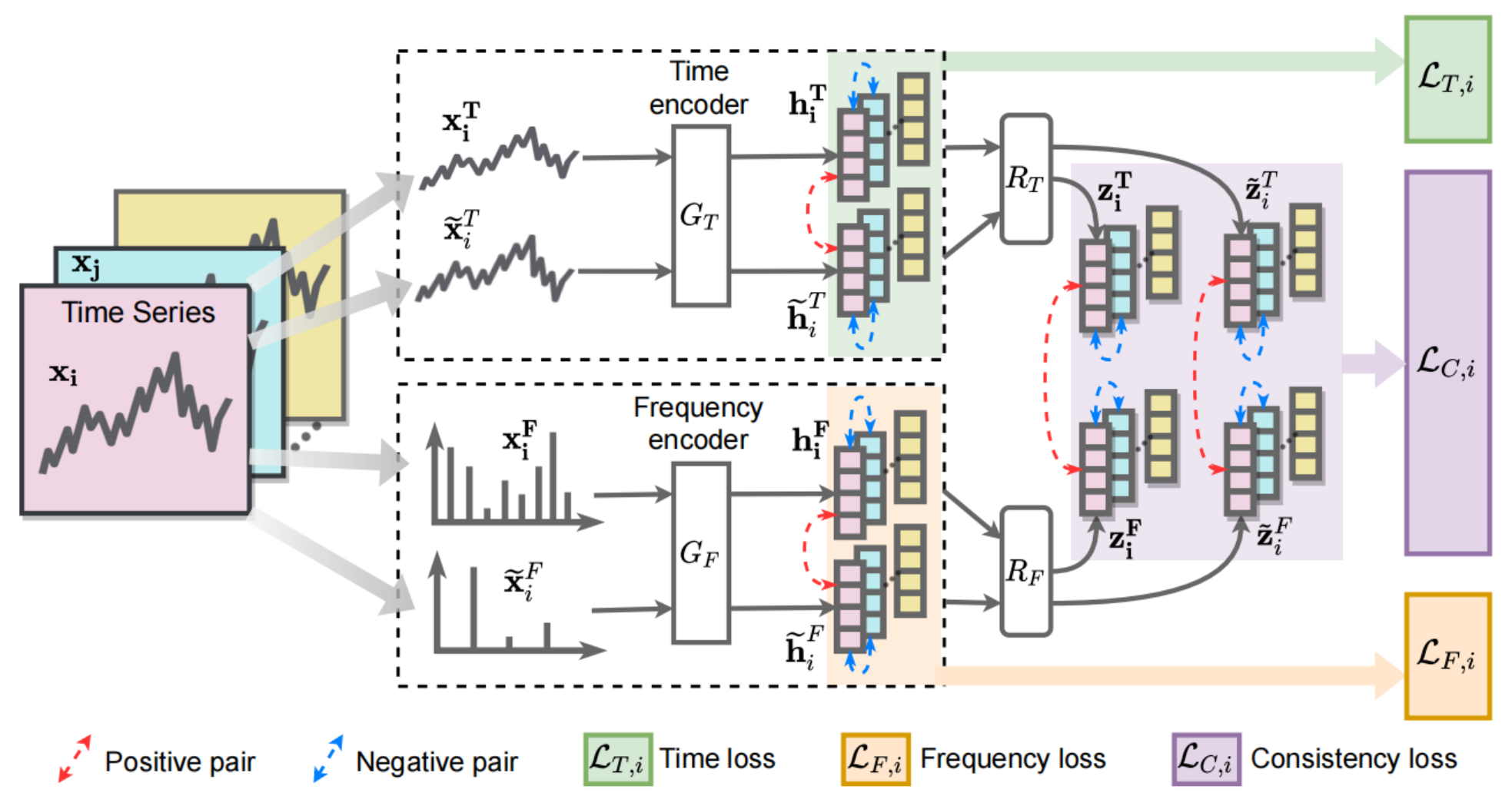}
\caption{\hspace{1em}The overview of TF-C}
\vspace{-0.1in}
\label{fig12}
\end{figure}
\noindent
\textbf{Contrastive pre-training.} Due to potential mismatches between pre-training and the target domain, there is a unique challenge in time series pre-training that may lead to diminished downstream performance. While domain adaptation methods can alleviate these changes \citep{berthelot2022adamatch,singh2021clda}, most approaches are considered suboptimal for pre-training as they often require direct examples from the target domain. To address this, these methods need to adapt to the diverse temporal dynamics of the target domain without relying on any target examples during pre-training.

Contrastive learning, a form of self-supervised learning, aims to train an input encoder to map positive sample pairs closer and negative pairs apart \citep{oord2018representation}. In time series, if the representations based on time and frequency for the same instance are close in the time-frequency space, it suggests a certain similarity or consistency in their features or attributes. \cite{zhang2022self}. proposed the need for Time-Frequency Consistency (TF-C) in pre-training, which involves embedding the time-based neighborhood of an example close to its frequency-based neighborhood. This work employs frequency-based contrastive enhancement to leverage rich spectral information and explore time-frequency consistency in time series. Contrastive pre-training can provide robust feature representations for forecasting tasks, contributing to enhanced model performance and generalization.\par

\begin{figure}[!t]
\centering
\setlength{\abovecaptionskip}{3mm}
\captionsetup{labelsep=none} 
\includegraphics[width=\linewidth]{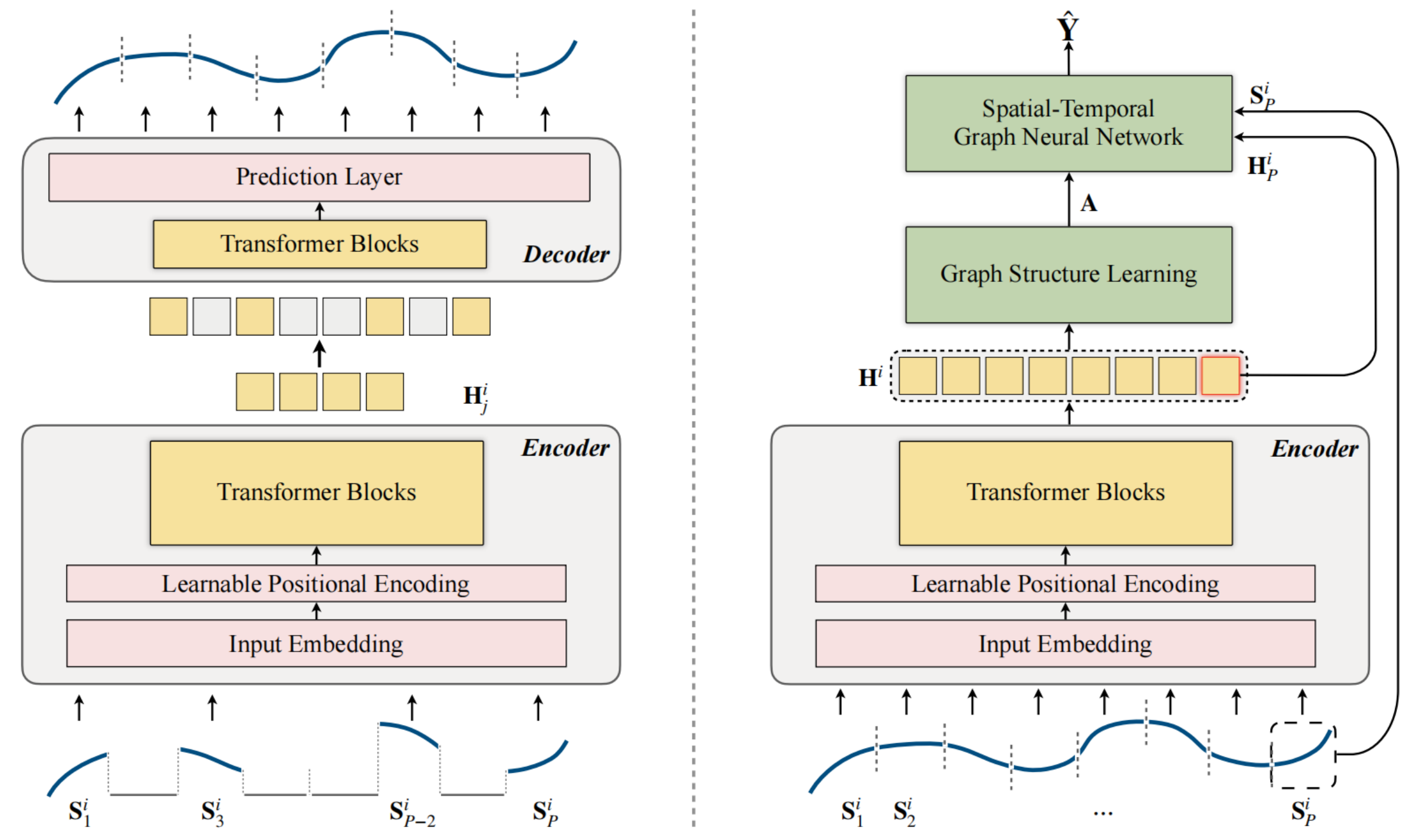}
\caption{\hspace{1em}The overview of STEP}
\vspace{-0.1in}
\label{fig13}
\end{figure}
\noindent
\textbf{Masking Pre-training.} Time series data is often continuous, ordered, but practically exhibits incompleteness. Additionally, real-world time series data commonly contains noise and uncertainty, necessitating models to possess robustness in dealing with such uncertainties. To address these crucial challenges in practice, the masking mechanism is regarded in some studies as an effective approach to enhance feature extraction.

In the work STEP, \cite{shao2022pre} designed an unsupervised pre-training model for time series based on Transformer blocks. The model employs a masked autoencoding strategy for training, which effectively learns temporal patterns and generates segment-level representations. These representations provide contextual information for subsequent inputs, facilitating the modeling of dependencies between short-term time series. The Ti-MAE model \citep{li2023ti} exhibits analogous efficacy in this regard. In the pre-training model SimMTM, \cite{dong2023simmtm} highlighted that randomly masking parts of the data severely disrupts temporal variations. They relate masking modeling to manifold learning and propose a Simple pre-training framework for Masked Time-series Modeling.

In summary, Masking pre-training simulates incompleteness and noise by masking some data points, enabling the model to learn how to handle partially missing information during the pretraining phase. This methodology can enhance the model's ability to capture long-term dependencies, increase tolerance to data uncertainty, and improve overall generalization performance.\par

\subsubsection{Patch-based segmentation}
In recent DTSF works, especially those of the Transformer models, the adoption of patch-based data organization has become prevalent \citep{nie2022time, lin2023petformer, zhang2023multi, ekambaram2023tsmixer, xue2023make, gong2023patchmixer}. It is advantageous to enhance the model's local perception capabilities by employing a patch-based strategy. Through segmenting long time series into smaller patches, the model becomes more adept at capturing short-term and local patterns within the sequence, thereby augmenting its comprehension of complex dynamics in the sequence. Simultaneously, the relationships among multivariate variables can yield information gain. Challenges lie primarily in how to learn the relationships among individual variables and introduce valid information into the model, while avoiding redundant information that may interfere with the model training process.

\begin{figure}[ht]
\centering
\setlength{\abovecaptionskip}{3mm}
\captionsetup{labelsep=none} 
\includegraphics[width=\linewidth]{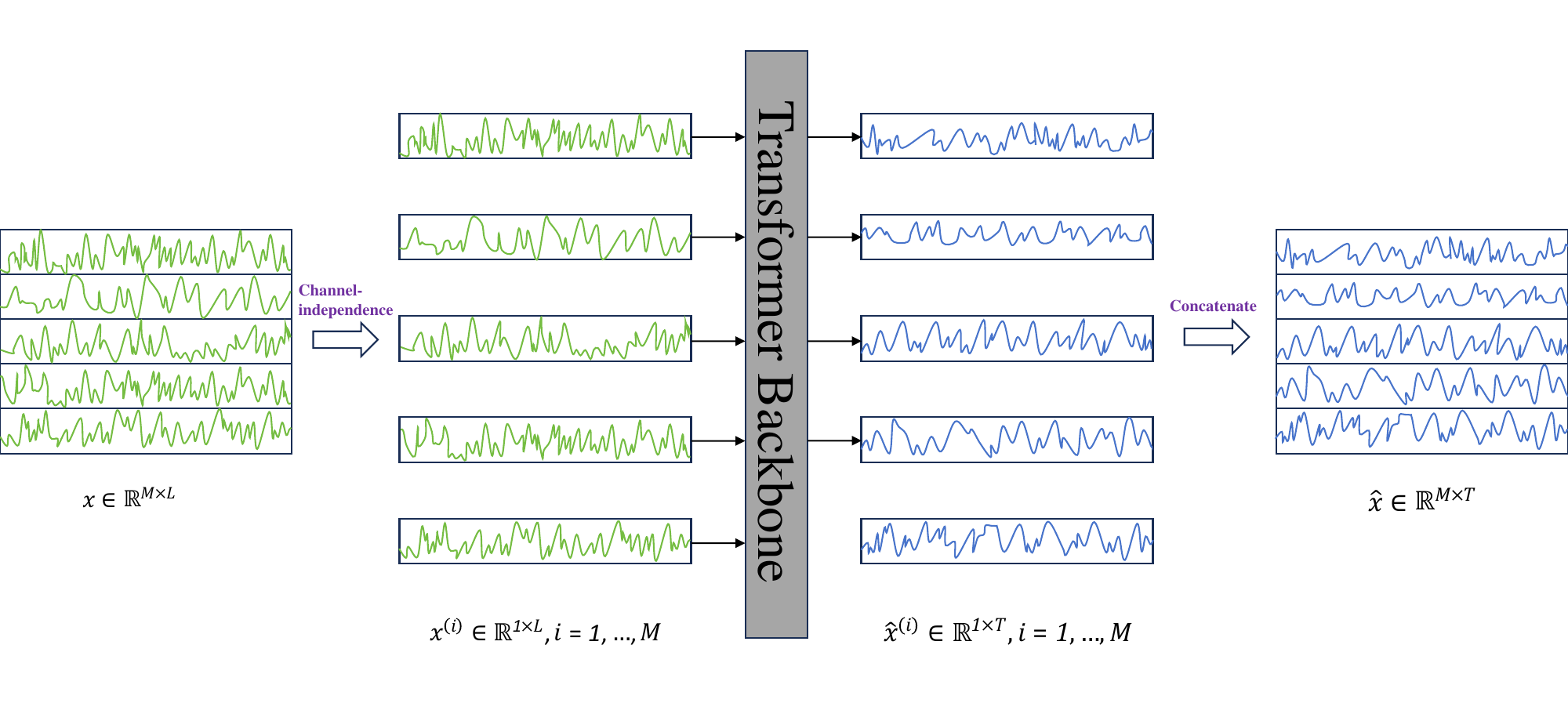}
\caption{\hspace{1em}The overview of PatchTST}
\vspace{-0.1in}
\label{fig14}
\end{figure}

\cite{nie2022time} proposed the PatchTST model, where they segment time series into subseries-level patches, serving as input tokens for the Transformer. They independently model each channel to represent a single variable. This channel-independent approach not only effectively preserves local semantic information for each variable in the embedding but also focuses on a more extended history. Furthermore, leveraging the channel-independent characteristics, potential feature correlations between single variables can be further learned through graph modeling methods \citep{zhang2023sageformer}. It allows for spatial aggregation of representations for global tokens in the graph. 

While the modeling emphasis varies across different works, there is a common consideration of employing methods that utilize subseries-level patches to process the raw time series data. This approach proves highly beneficial for capturing and learning the local features of the data. The patch-based segmentation method introduces another methodology for TSF. Additionally, channel independence emerges as a viable avenue for exploring multivariate time series forecasting.

\begin{table}[ht]
    \renewcommand{\arraystretch}{1.2} 
    \footnotesize
    \captionsetup{labelsep=none} 
    \caption{\hspace{1em} Time Series Datasets in Primary Domains. The table summarizes commonly used datasets and indicates whether they are multivariate, which implies temporal alignment with known timestamps}
    \vspace{0.03in}
    \label{Datasets_m}
    \begin{tabularx}{\columnwidth}{
        >{\centering\arraybackslash}m{2cm}|
        >{\centering\arraybackslash}m{1.5cm}
        >{\centering\arraybackslash}m{1cm}
        >{\centering\arraybackslash}m{1.5cm}
        >{\centering\arraybackslash}m{1.5cm}
        >{\centering\arraybackslash}m{1cm}
        >{\centering\arraybackslash}m{1.5cm}
        >{\raggedright\arraybackslash}p{3.5cm}} 
        \hline

        Domain & Datasets & Variants & Data Time Range & Data Granularity & Multi/Uni & Authors \\
        \midrule[0.8pt]
        &\href{https://github.com/zhouhaoyi/ETDataset} {ETTh1} & 7 & 2016 - 2018  & 1h & Multi+Uni & \makecell{Zhou\\et al.}   \\
        
        & \href{https://github.com/zhouhaoyi/ETDataset}{ETTm1} & 7 & 2016 - 2018 & 15m & Multi+Uni & \makecell{Zhou\\et al.}    \\
        
        \makecell[c]{Energy} & \href{https://archive.ics.uci.edu/ml/datasets/ElectricityLoadDiagrams20112014}{Electricity} & 321 & 2011 - 2014 & 1h & Multi+Uni & -   \\
        
        & \href{https://www.kaggle.com/datasets/sohier/30-years-of-european-wind-generation}{Wind} & 28 & 1986 - 2015 & 1h & Uni & -  \\
        
        & \href{http://www.nrel.gov/grid/solar-power-data.html}{Solar-Energy} & 137 & 2006 - 2006 & 10m & Multi+Uni & Solar   \\
        \midrule[0.8pt]
        
        \makecell{\\\\Healthcare} & \href{https://gis.cdc.gov/grasp/fluview/fluportaldashboard.html}{ILI} & 7 & 2002 - 2021 & 1w & Uni & -   \\
        
        & \href{http://ecg.mit.edu/}{MIT-BIH} & 2 & 1975 - 1979 & 360Hz & Uni & George   \\
        \midrule[0.8pt]
        
        & \href{https://pems.dot.ca.gov}{Traffic} & 862 & 2015 - 2016 & 1h & Uni & Caltrans   \\
        
        \raisebox{+3.0ex}{Transportation} & \makecell{\href{https://journals.sagepub.com/doi/abs/10.3141/1748-12}{PeMSD4}\\\href{https://journals.sagepub.com/doi/abs/10.3141/1748-12}{PeMSD7}\\\href{https://journals.sagepub.com/doi/abs/10.3141/1748-12}{PeMSD8}} & \makecell{307\\228\\170} & \makecell{2018/1\\2012/5\\2016/7} & 5m & Multi & \makecell{Chen\\et al.}  \\
        \midrule[0.8pt]
          
        &\href{https://www.ncei.noaa.gov/data/local-climatological-data/} {Weather1} & 12 & 1981 - 2010 & 1h & Uni & -   \\
        
        \makecell[c]{Meteorology} & \href{https://www.bgc-jena.mpg.de/wetter/}{Weather2} & 21 & 2020 - 2021 & 10m & Multi+Uni & \makecell{Sparks\\et al.}   \\
        
        &\href{https://zenodo.org/records/5129091}{ \makecell[c]{Temperature\\Rain}} & 2 & 2015 - 2017 & 1d & Multi+Uni & \makecell{Rakshitha\\et al.}   \\
        \midrule[0.8pt]
        
        & \href{https://datahub.io/core/exchange-rates}{Exchange-Rate} & 8 & 1990 - 2016 & 1d & Uni & Lai et al.   \\
        
        \makecell[c]{Economics} & \href{http://urn.fi/urn:nbn:fi:csc-kata20170601153214969115}{LOB-ITCH} & 149 & 2010 - 2010 & 1ms-10min & Uni & \makecell{Adamantios\\et al.}   \\
        
        & \href{https://zenodo.org/records/4654802}{Dominick} & 25 & 1989 - 1994 & 1w & Uni & \makecell{Godahewa\\et al.}   \\
        \bottomrule[1.5pt]
    \end{tabularx}
\end{table}

\section{Challenges and Prospects}\label{sec6}
\begin{figure*}[!t]
\centering
\setlength{\abovecaptionskip}{3mm}
\captionsetup{labelsep=none} 
\includegraphics[width=\linewidth,height=\textheight,keepaspectratio]{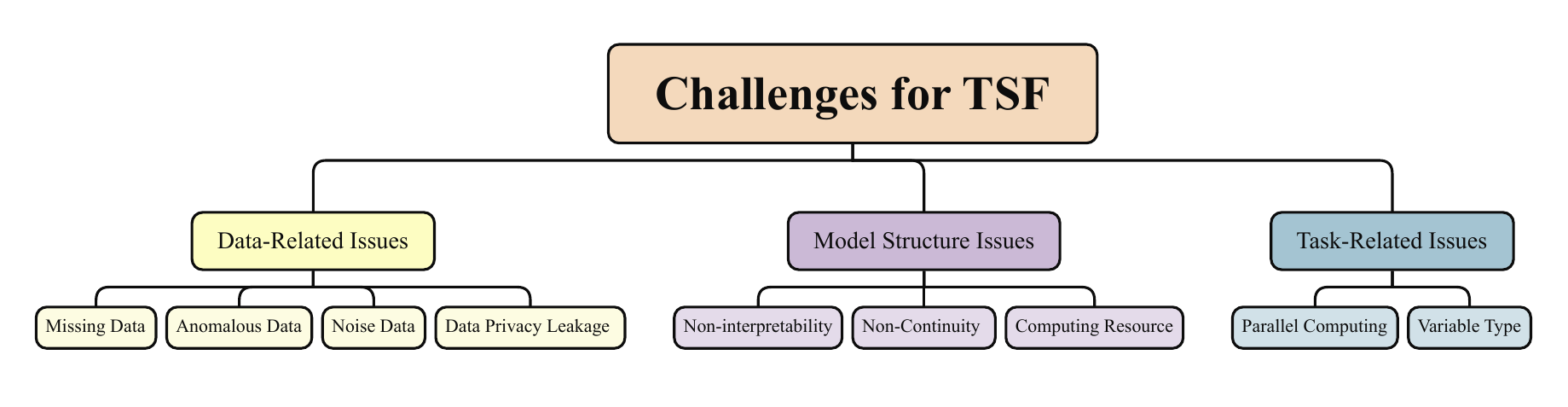}
\caption{ \hspace{1em}Challenges in Time Series Forecasting}
\vspace{-0.1in}
\label{fig_cha_2}
\end{figure*}

We have investigated the neural network architectures, feature extraction and learning approaches, and significant experimental datasets of deep learning models in the context of TSF. While DTSF models have demonstrated remarkable achievements across diverse domains in recent years, certain challenging issues remain to be addressed, which point towards potential future research directions. We summarize these challenges and propose viable avenues as follows. We classify the challenges into three main categories: data features, model structure, and task-related issues. Within each category, we highlight several representative challenges. Figure \ref{fig_cha_2} illustrates an overview of these challenges.


\subsection{Challenges}
\subsubsection{Lack of Data Privacy Protection and Completeness} Federated learning (FL) is gaining momentum in the field of TSF, primarily addressing challenges associated with large local data volumes and privacy concerns during information exchange. With FL, multiple participants can collaboratively train models without the need to share sensitive raw data \citep{mcmahan2017communication}. In TSF tasks, each participant can leverage their local time series data for model training. Through FL algorithms, the parameters of local models are aggregated to obtain a global predictive model. This distributed learning process ensures privacy protection, mitigating the risks of privacy breaches associated with centralized data storage and transmission. Current research efforts predominantly focus on load detection \citep{gao2021decentralized, taik2020electrical, briggs2022federated}, traffic speed and flow \citep{zhang2021fastgnn,liu2020privacy}, energy consumption \citep{zhang2020probabilistic,savi2021short}, and communication networks \citep{subramanya2021centralized,diaz2019federated}, among others. Exploring feasible solutions in other domains remains an open avenue. Furthermore, federated learning harnesses the diversity of distributed data sources, thereby enhancing model generalization and prediction accuracy. Hence, federated learning holds great promise in the realm of TSF, offering a prospective solution for large-scale, secure, and efficient time series prediction and analysis.\par
 
\subsubsection{Lack of Interpretability} So far, the majority of efforts in the field of TSF have primarily focused on enhancing predictive performance through the design of intricate model architectures. However, research into the interpretability of these models has been relatively limited. As neural networks find application in critical tasks \citep{moraffah2020causal}, the demand for comprehending why and how models make specific predictions has been growing. The N-BEATS model achieves high accuracy and interpretability in TSF by designing the interpretable architecture and output mechanisms \citep{oreshkin2019n}. This enables users to better comprehend the model's predictive outcomes while maintaining high forecasting precision.

Post-hoc interpretable models are developed for the purpose of elucidating already trained networks, aiding in the identification of crucial features or instances without modifying the original model weights. These approaches mainly fall into two categories. One involves the application of simpler interpretable surrogate models between the inputs and outputs of the neural network, relying on these approximate models to provide explanations \citep{ribeiro2016should,lundberg2017unified}. The other category encompasses gradient-based methods, such as those presented in \citep{simonyan2013deep, koh2017understanding, siddiqui2019tsviz}, which scrutinize the network gradients to determine which input features exert the most significant influence on the loss function.

Furthermore, it is noteworthy that, in contrast to the black-box nature of traditional neural networks, a series of TSF models based on the Transformer architecture incorporate attention layers with inherent interpretability. These attention layers can be strategically integrated into other models, with the analysis of attention weights aiding in the comprehension of the relative importance of features at each time step \citep{choi2016retain,li2019enhancing,bai2018interpretable}. By scrutinizing the distribution of attention vectors across time intervals, the model can gain better insights into persistent patterns or relationships within the time series \citep{lim2021temporal}, such as seasonal patterns.

Recent advancements in the field have focused on learning from perturbations and interpretable sparse system identification methods to enhance the interpretability of time series data \citep{enguehard2023learning, liuinterpretable}. Among these, sparse optimization methods, which obviate the need for time-consuming backpropagation training, exhibit efficient training capabilities on CPUs. These methods offer insights for further exploration into interpretable time series forecasting.\par

\subsubsection{Lack of Temporal Continuity} Compared to traditional deep learning forecasting models, the proposal of the Neural Ordinary Differential Equation (NODE) \citep{chen2018neural} has directed our attention towards the derivatives of neural network parameterized hidden states, which showcases superior performance over RNNs in both continuous and discrete time series problems. Recent studies applying Ordinary Differential Equations (ODE) or Partial Differential Equations (PDE) to TSF have explored various directions such as learning latent relationships between variables or events \citep{li2022learning, de2019gru, gao2022egpde}, handling irregular data \citep{scholz2022latent}, achieving interpretable continuity \citep{gao2022explainable,jin2022multivariate}, optimizing model parameters \citep{chen2011time}, and exploring differential dynamics \citep{linot2023stabilized,gilani2021diffusion}. The ETN-ODE model proposed by \cite{gao2022explainable} is the first interpretable continuous neural network for multi-step time series forecasting of multiple variables at arbitrary time instances. Additionally, their EgPDE-Net model \citep{gao2022egpde} is also the first to establish the continuous-time representation of multivariate time series as a partial differential equation problem. Its specially designed architecture utilizes ODE solvers to transform the partial differential equation problem into an ODE problem, facilitating predictions at arbitrary time steps.

Temporal continuation is one of the crucial factors to consider in the TSF process. The application of the Neural Differential Equation (NDE) paradigm in DTSF integrates DL with differential equation modeling to naturally and accurately capture the dynamic evolution of time series. It interprets the evolution of individual components more clearly and flexibly captures instantaneous changes by using a differential equation to describe the rate of change of the data at each time point. For deep learning modelling of complicated time series data, the NDE technique offers an innovative and effective paradigm.\par

\subsubsection{Challenges of Parallel Computing} In the era of massive data, there is an urgent demand for online real-time analysis of time series data. Currently, time series models are constructed based on stand-alone sequence analysis, which often requires the use of high-performance GPU servers to improve computational efficiency. However, on one hand, it is constrained by computational resources and data scale, making real-time online forecasting unattainable. On the other hand, GPU servers are costly. Therefore, the research on efficient parallel computing based on deep learning and big data analytics technologies is poised to become a critical challenge.\par

\subsubsection{Challenges of Large Models} 
Large models demonstrate advantages in the field of time series forecasting, excelling in capturing long-term dependencies, handling high-dimensional data, and mitigating noise. A noteworthy exploration in this direction occurred on December 13, 2023 when Amazon released work utilizing large models for time series forecasting, marking a pioneering effort in applying large models to temporal prediction \citep{xue2023promptcast}. This work leverages large models to construct intricate relationships between sequences while harnessing their robust text data processing capabilities. The integration of large models has enhanced the handling of multimodal data and interpretability in financial forecasting scenarios. Large models have already ventured into various domains, encompassing stock price predictions in financial markets \citep{zhou2023one, jin2023time, chang2023llm4ts}, inference of medical data \citep{sun2023test,gruver2023large}, forecasting human mobility trajectories \citep{cao2023tempo}, and serving as general-purpose models for weather and energy demand predictions \citep{yu2023temporal, xie2023wall, zhang2023instruct, liu2023large, li2023frozen}.

On another note, significant strides have been made in the training of foundational time series models \citep{xue2022leveraging,garza2023timegpt}. The recent TimeGPT-1 model \citep{rasul2023lag} applies the techniques and architecture underlying large language models (LLM) to the forecasting domain, successfully establishing the first foundational time series model capable of zero-shot inference. This breakthrough opens avenues for creating foundational models specifically tailored for time series forecasting.

We believe that the performance and value of large models in the realm of time series forecasting will continue to unfold as technological advancements and innovations progress.
\par

\subsection{Prospects}
\subsubsection{Potential Representation Learning} Representation Learning (RL) has recently emerged as one of the hot topics in time series forecasting. While models based on stacked layers can yield respectable results, they often come with high computational costs and may struggle to capture the inherent features of the data. RL, on the other hand, focuses on acquiring meaningful latent features that result in lower-dimensional and compact data representations, capturing the fundamental characteristics of the data. Presently, many self-supervised or unsupervised approaches aim to encode raw sequences to learn these latent representation features \citep{eldele2023self,darban2023carla}. Some works employ multi-module architectures or model ensembles \citep{mehrkanoon2019deep, lyu2018improving, yang2019representation}, while others use pre-training with denoising, smoothing properties, siamese structures or 2D-variation modeling \citep{zheng2023simts,zerveas2021transformer, wu2022timesnet}, which provide novel solutions to various domain-specific problems. Besides, contrastive learning is dedicated to enabling models to compare observations at different time points and learn rich data representations by contrasting positive and negative samples. Some works \citep{yue2022ts2vec,zhang2022self,ozyurt2022contrastive,luo2023time} have utilized contrastive learning to assist models in learning meaningful features from unlabeled data, thus enhancing their generalization performance. This is especially valuable when labeled data is limited or unavailable.

Learning temporal representations and employing contrastive training can significantly enhance the model's representation and generalization capabilities in TSF. This greatly improves the model's performance in handling complex, noisy, or changing data distributions.\par

\subsubsection{Counterfactual Forecast and Causal Inference} Counterfactual forecasting and causal inference represent promising avenues for future research in DTSF. Despite the existence of lots of deep learning methods for estimating causal effects in static settings \citep{yoon2018ganite, hartford2017deep, alaa2017deep}, the primary challenge in time series data lies in the presence of time-dependent confounding effects. This challenge arises due to the time-dependence, where actions that influence the target are also conditioned on observations of the target. Recent research advancements encompass the utilization of statistical techniques, novel loss functions, extensions of existing methods, and appropriate inference algorithms \citep{lim2018forecasting, bica2020crn, li2020g, liu2023causal, gao2023causal}. 

Moreover, while some efforts provide counterfactual explanations for time series models \citep{dhaou2021causal, nemirovsky2022countergan}, they fall short of generating realistic counterfactual explanations or feasible counterfactual explanations for time series models. Recent work has introduced a self-interpretable model capable of generating actionable counterfactual explanations for time series forecasting \citep{yan2023self}.

Future research directions may revolve around further refining these approaches to address the additional complexities inherent in time series data and get more accurate counterfactual interpretations. Additionally, innovative methods should be sought to harness the full potential of deep learning in counterfactual forecasting and causal inference, ultimately enhancing decision-making processes across various domains.\par

\subsubsection{TS Diffusion} The burgeoning development of Diffusion models in the domain of image and video streams has sparked novel theories and models, gradually extending into the realm of TSF. Notably, TimeGrad employs RNN-guided denoising for autoregressive predictions \citep{rasul2021autoregressive}, while CSDI utilizes non-autoregressive methods with self-supervised masking \citep{NEURIPS2021_cfe8504b}. Similarly, SSSD utilizes structured state-space models to reduce computational complexity \citep{alcaraz2022diffusion}. Despite being early explorations in the TSF domain, these models still suffer from slow inference, high complexity, and boundary inconsistencies.

In recent researches, the unconditionally trained TSDiff model employs self-guidance mechanisms to alleviate the computational overhead in reverse diffusion for downstream task forecasting without auxiliary networks \citep{kollovieh2023predict}. TimeDiff addresses boundary inconsistencies with future mixups and autoregressive initialization mechanisms \citep{shen2023non}. The multi-scale diffusion model MR-Diff leverages multi-resolution temporal structures for sequential trend extraction and non-autoregressive denoising \citep{shen2024multi}.

The first framework based on DDPM, Diffusion-TS, accurately reconstructs samples using Fourier-based loss functions, extending to forecasting tasks \citep{yuan2024diffusionts}. Furthermore, the TMDM model combines conditional diffusion generation processes with Transformer to achieve precise distribution prediction for multivariate time series \citep{litransformer}.

The work on Diffusion primarily focuses on denoising, and numerous groundbreaking initiatives are emerging in the realm of DTSF. We anticipate Diffusion to become a prominent direction.

\subsubsection{Determine the Weight of the Aggregate Model}
At present, ensemble learning, as one of the mainstream paradigms, has proven to be effective and robust \citep{taylor2009wind, makridakis2018statistical, arbib2003handbook}. However, determining the weights of base models in an ensemble remains an unsolved challenge. Sub-optimal weighting can hinder the full potential of the final model. To address this challenge, \cite{fu2022reinforcement} proposed a model combination framework based on reinforcement learning (RLMC). It uses deterministic policies to output dynamic model weights for non-stationary time series data and leverages deep learning to extract hidden features from raw time series data, allowing rapid adaptation to evolving data distributions. Notably, in RLMC, the use of DDPG, an off-policy actor-critic algorithm \citep{lillicrap2015continuous}, can produce continuous actions suitable for model combination problems and is trained with recorded data to achieve improved sample efficiency. Therefore, the combination of reinforcement learning with some continuous control algorithms \citep{fujimoto2018addressing,haarnoja2018soft} presents a unique utility in determining ensemble model weights and is a path worth exploring.\par

\subsubsection{Interdisciplinary Exploration} Due to the multidimensional nature of the relationships between causes and effects in reality, there exist complex interconnections among time series. While deep learning models have demonstrated excellent performance in tackling intricate TSF problems, they often lack systematic interpretability and clear hierarchical structures. In the realm of network science, when dealing with extensive data, numerous variables, and intricate interconnections, it is possible to construct multi-layered networks by categorizing and stratifying the relationships among various elements. By examining the dynamic changes in multi-layered networks, it becomes feasible to forecast multidimensional data by analyzing high-dimensional correlations.

For diverse domains, an interdisciplinary approach, such as incorporating network science or other relevant theories, can be a beneficial choice in the future of DTSF research. This approach enables a more insightful analysis of problems and their multidimensional aspects.

\section{Conclusion}\label{sec7}

In this paper, we present a systematic survey for deep learning-based time series forecasting. We commence with the fundamental definition of time series and forecasting tasks and summarize the statistical methods and their shortcomings. Next, moving on, we delve into neural network architectures for time series forecasting, summarizing five major model paradigms that have gained prominence in recent years: the Encoder-Decoder, Transformer, Generative Adversarial, Integration, and Cascade. Furthermore, we conduct an in-depth analysis of time series composition, elucidating the primary approaches to enhance feature extraction and learning from time series data. Additionally, we survey time series forecasting datasets across major domains, encompassing energy, healthcare, traffic, meteorology, and economics. Finally, we comprehensively outline the current challenges in the field and propose some potential research directions.

\section*{Declarations}

\begin{itemize}
\item{\textbf{Funding}: This work was supported in part by the National Natural Science Foundation of China under Grant 62476247, 6207395 and 62072409, in part by the \verb+"+Pioneer\verb+"+ and \verb+"+Leading Goose\verb+"+ R\&D Program of Zhejiang under Grant 2024C01214, and in part by the Zhejiang Provincial Natural Science Foundation under Grant LR21F020003.}

\item{\textbf{Confict of interest}: The authors declare that they have no known competing financial interests or personal relationships that could have appeared to influence the work reported in this paper.}
\item{\textbf{Data availability}: Data sharing is not applicable to this article as no new data were created or analyzed in this study. }

\end{itemize}


\bibliographystyle{plainnat} %
\bibliography{sn-article}

\appendix
\newpage
\section{Datasets in Different Domain}\label{sec5}

Time series, which exists in every aspect of our lives, carries the historical data of various fields in the time dimension. Many datasets have been accumulated during the development of the TSF task. These datasets are often cited in top conferences and journals within the computer domain, furnishing researchers with high-quality research data characterized by rich samples and features, thus holding significant reference value. However, the diversity of these datasets introduces a significant challenge—data heterogeneity. The datasets described below cover five key TSF application areas: energy, transportation, economics, meteorology, and healthcare \citep{gebodh2021dataset}, as shown in Table \ref{Datasets_m}. These fields feature data with varying structures, formats, time granularities, and scales, such as sensor data, text, and images, complicating model construction. To address these issues, several techniques have been proposed.

Multimodal learning, through shared representation learning, integrates diverse data types, improving model handling of heterogeneous data \citep{guo2019deep}. Time alignment techniques, such as the TAM model, synchronize data from different time granularities by introducing a novel time-distance measure \citep{folgado2018time}. Deep generative models, like GinAR, address missing values and noise by generating new samples and rebuilding spatiotemporal dependencies \citep{yu2024ginar}. Self-supervised learning methods, such as SimCLR, allow models to learn from unlabeled data, improving adaptability to heterogeneous sources \citep{chen2020simple}. Finally, collaborative attention mechanisms capture complex correlations between multimodal data and adjust modality weights dynamically, enhancing model learning capacity \citep{dosovitskiy6909discriminative}. These models and techniques effectively integrate heterogeneous data, improving the stability and accuracy of time series forecasting in multi-source environments.

\subsection{Energy}
TSF is currently being extensively applied in a prominent domain, namely, energy management. Accurate forecasting within this domain plays a crucial role in facilitating status assessment and trend analysis, which in turn enables the implementation of intelligent strategies in engineering planning. Fortunately, modern energy systems autonomously gather extensive datasets encompassing diverse energy sources such as electricity \citep{singh2018big}, wind energy \citep{feng2017characterizing}, and solar energy \citep{rajagukguk2020review}. These data resources are leveraged for the identification of patterns and trends in energy demand and supply, providing valuable insights for the development of advanced forecasting models.

\begin{figure}[!t]
\centering
\setlength{\abovecaptionskip}{3mm}
\captionsetup{labelsep=none} 
\includegraphics[height=10.5cm,width=\linewidth]{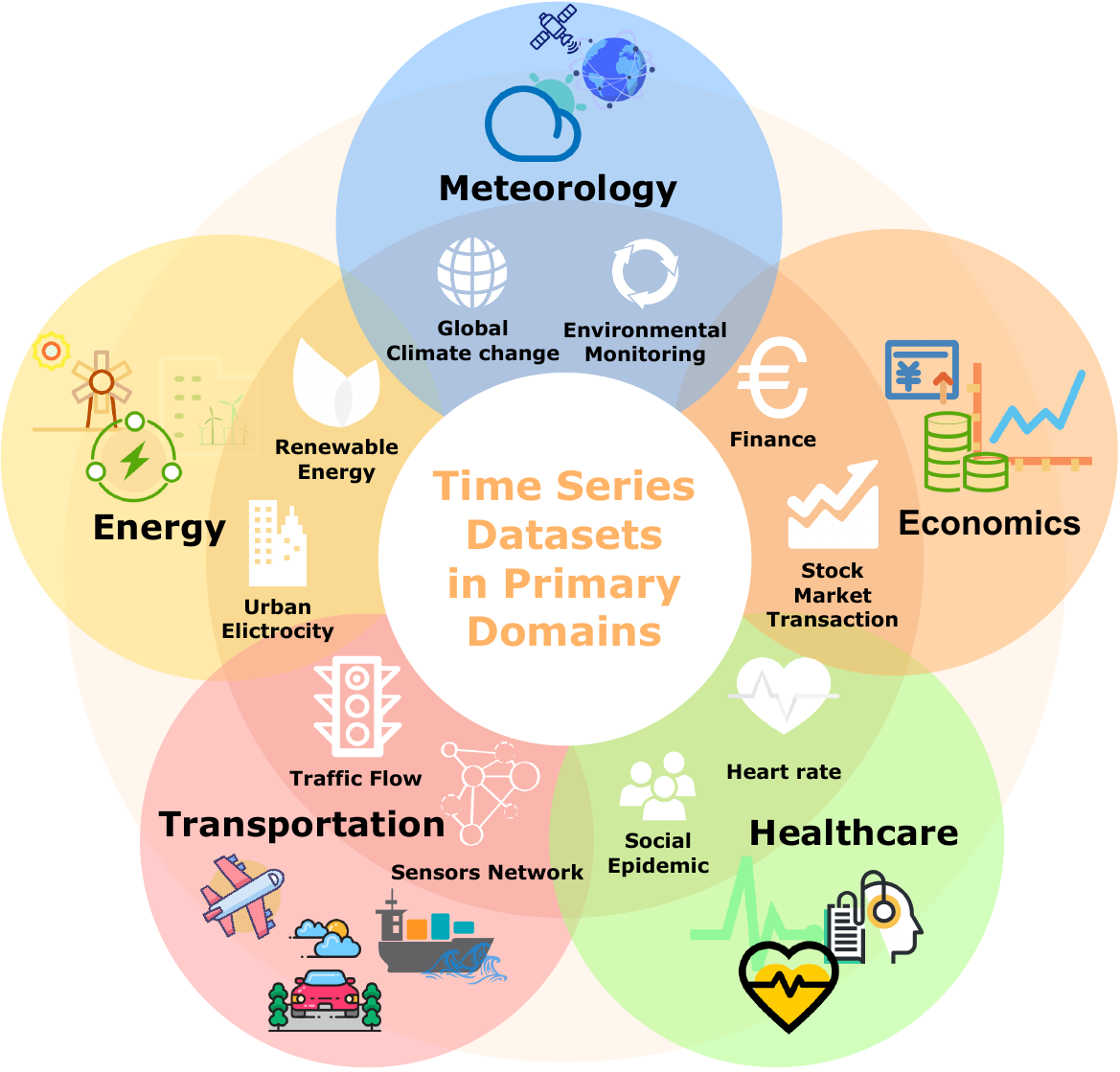}
\caption{\hspace{1em}Time series datasets in primary domains}
\vspace{-0.1in}
\label{fig15}
\end{figure}

\subsubsection{Electricity Transformer Temperature (ETT)} The ETT-small dataset encompasses data originating from two distinct power transformer installations, each situated at a separate site \citep{zhou2021informer}. This dataset comprises a variety of parameters, such as load profiles and oil temperature readings. It serves the purpose of predicting the oil temperature of power transformers and investigating their resilience under extreme load conditions. The temporal scope of this dataset spans from July 2016 to July 2018, with data recorded at 15-minute intervals. These datasets originate from two geographically disparate regions within the same province in China, designated as ETT-small-m1 and ETT-small-m2, respectively. Each of these datasets consists of an extensive 70,080 data points, calculated based on a duration of 2 years, 365 days per year, 24 hours per day, and data sampling at 15-minute intervals. Furthermore, the dataset offers an alternate version with hourly granularity, denoted as ETT-small-h1 and ETT-small-h2. Each data point within the ETT dataset is characterized by an 8-dimensional feature vector, which includes the timestamp of the data point, the target variable 'oil temperature', and six distinct types of external load values.

\subsubsection{Electricity} The initial dataset utilized in this investigation is the Electricity Load Diagrams 2011-2014 Dataset \citep{misc_electricityloaddiagrams20112014_321}, which records 370 customers' electricity usage information between 2011 and 2014. Data is recorded in the original dataset every 15 minutes. It was necessary to preprocess the dataset by deleting the 2011 data and aggregating it into hourly consumption in order to address the problem of some dimensions having a value of 0. As a result, the final dataset includes information on 321 customers' electrical use from 2012 to 2014.
\subsubsection{Wind (European Wind Generation)} For 28 European countries between 1986 and 2015, this dataset \footnote{https://www.kaggle.com/datasets/sohier/30-years-of-european-wind-generation} offers hourly estimates of energy potential expressed as a percentage of the maximum output from power plants. It is distinguished from other datasets by having sparser data and a notable frequency of zeros at regular intervals.
\subsubsection{Solar-Energy} The solar power production of 137 photovoltaic plants in Alabama State in 2006, recorded at 10-minute intervals, constitutes the dataset for our evaluation of short-sequence forecasting capabilities \footnote{https://www.nrel.gov/grid/solar-power-data.html}.

\subsection{Healthcare}
TSF plays a pivotal role in the healthcare domain, serving as a critical tool for predicting disease onset and progression, evaluating the efficacy of pharmaceutical interventions, and monitoring fluctuations in patients' vital signs. These forecasts empower healthcare practitioners in enhancing disease diagnosis, devising treatment strategies, overseeing patient well-being, and implementing preventive measures for disease surveillance and containment.

\subsubsection{ILI (Influenza-Like Illness)} Weekly reports from the US Centers for Disease Control and Prevention from 2002 to 2021 are included in the set of data. It contains data on the overall number of patients as well as the percentage of patients having influenza-like symptoms.
\subsubsection{EEG (Electroencephalogram)} The collection includes EEG \footnote{https://github.com/meagmohit/EEG-Datasets} recordings of participants obtained both prior to and during the performance of mental math exercises. Every recording is made up of 60-second EEG segments free of artifacts. For every subject in the dataset, there are 36 CSV files total, and each file has 19 data channels. 

\subsubsection{MIT-BIH (Arrhythmia Database)} There are 48 half-hour segments of two-channel ambulatory ECG recordings available in the MIT-BIH Arrhythmia Database \footnote{http://ecg.mit.edu/}. These recordings were from 47 individuals that the BIH Arrhythmia Laboratory examined from 1975 to 1979. Every recording was digitalized with a resolution of 11 bits and a range of 10 mV, at a rate of 360 samples per second per channel. Electrocardiogram data from this dataset can be used for anticipating arrhythmias, among other uses.
\par

\subsection{Transportation}
Accurate and timely TSF of traffic is vital for urban traffic control and management. It aids in predicting traffic congestion, traffic flow, accident rates, and the utilization of public transportation. These predictions can be used by transportation authorities and companies to plan and manage transportation systems more effectively, thereby improving traffic efficiency and safety.

\subsubsection{Traffic} This dataset \footnote{https://pems.dot.ca.gov/} includes hourly data from 2015–2016 that was collected during a 48-month period from the California Department of Transportation. The statistic shows the hourly road occupancy rate, which ranges from 0 to 1. The San Francisco Bay Area's roadways are home to 862 different sensors from which the measurements are obtained.
\subsubsection{PeMSD4/7/8}
These datasets are highly regarded as industry standards for traffic forecasting \citep{chen2001freeway}. 

PeMSD4 is one of them and it includes traffic speed data from the San Francisco Bay Area. It incorporates data from 29 roads' worth of 307 sensors. The January–February 2018 time frame is covered by the dataset.

PeMSD7 includes traffic information from California's District 7. It covers the workday period from May to June 2012 and includes traffic speeds recorded by 228 sensors. Five minutes are allotted for the collection of data.

PeMSD8 contains San Bernardino traffic statistics taken during July and August of 2016. It includes data from 170 detectors positioned along 8 distinct routes. Five minutes are allotted for the collection of data.
\par

\subsection{Meteorology}
TSF has become an indispensable task in the field of meteorology with wide-ranging applications in weather forecasting, such as meteorological disaster warnings, agricultural production, and more.

\subsubsection{Weather1}
The dataset Weather1 encompasses climate data from almost 1600 locations in the United States \footnote{https://www.ncei.noaa.gov/data/local-climatological-data/}, spanning a 4-year period from 2010 to 2013. Hourly data points were collected, featuring the target value \verb+"+wet bulb\verb+"+ and 11 climate-related features.

\subsubsection{Weather2} 
Weather2 comprises a meteorological time series featuring 21 weather indicators \footnote{https://www.bgc-jena.mpg.de/wetter/}, collected every 10 minutes in 2020 by the Max Planck Institute for Biogeochemistry's weather station.

\subsubsection{Temperature Rain}
Consisting of 32,072 daily time series, this dataset \citep{godahewa_2021_5129091} presents temperature observations and rain forecasts collected by the Australian Bureau of Meteorology. The data spans 422 weather stations across Australia, covering the period from 02/05/2015 to 26/04/2017.

\subsection{Economics}
In the field of finance, one of the most extensively studied areas in TSF is the prediction of financial time series, particularly asset prices. Typically, there are several subtopics in this field, including stock price prediction, index prediction, foreign exchange price prediction, commodity (such as oil, gold, etc.) price prediction, bond price prediction, volatility prediction, and cryptocurrency price prediction. The following section will introduce commonly used datasets in this domain.

\subsubsection{Exchange-Rate} This dataset \citep{lai2018modeling} compiles daily exchange rates mainly in trading days for eight countries (Australia, Canada, China, Japan, New Zealand, Singapore, Switzerland, and the United Kingdom) spanning the years 1990 to 2016.

\subsubsection{LOB-ITCH} Due to the lack of adequate records, few other fields have Millisecond data on the span of days as in finance. In the financial field, with the advent of automated trading, limit order books were born, which are very conducive to high-frequency traders' operations and leave a large amount of detailed data. The LOB-ITC dataset comprises around four million events, each with a 144-dimensional representation, pertaining over five stocks for ten consecutive trading days \citep{ntakaris2018benchmark}, from June 1, 2010 to June 14, 2010. And what makes this data different from other data of the same kind is the centralized trading market in the Nordic region.Some researchers found that  \verb+"+the differences between different trading platforms' matching rules and transaction costs complicate comparisons between different limit order books for the same asset \citep{o2011market}\verb+"+ . Therefore, Stock Exchange, which has decentralized exchanges like the United States, has more influencing factors and is more difficult to  model. In contrast, Helsinki Exchange is a pure limit order market, which can provide purer data.

\subsubsection{Dominick} This dataset \citep{godahewa_2021_4654802} incorporates data from randomized experiments conducted by the University of Chicago Booth School of Business and the now-defunct Dominick’s Finer Foods. The experiments spanned from 1989 to 1994, covering over 25 different categories across all 100 stores in the chain. As a result of this research collaboration, approximately nine years of store-level data on the sales of more than 3,500 UPCs are available through this resource.\par

\subsection{Further Data Sources}
In addition to the commonly used datasets mentioned above, we extensively surveyed data sources from various domains and compiled a subset of additional datasets. These datasets are derived from influential works and serve as the foundation for researching niche topics and detailed investigations in respective fields. We will provide appropriate descriptions of the datasets listed in Table \ref{further_data}.\par
Several comprehensive datasets from large-scale competitions are also noteworthy, such as M3/M4/M5. These datasets were put forward by the Makridakis Competitions, which are a series of open competitions to evaluate and compare the accuracy of different TSF methods.

\begin{table}[!h]
    \footnotesize
    \captionsetup{labelsep=none} 
    \caption{\hspace{1em}Summary of the datasets used in the experiments}
    \label{further_data}
    \renewcommand{\arraystretch}{0.5}
  \centering
  \begin{tabular}
    {>{\centering\arraybackslash}m{1.8cm}|
     >{\centering\arraybackslash}m{1.2cm}
     >{\centering\arraybackslash}m{3.0cm}
     >{\centering\arraybackslash}m{2.8cm}
     >{\centering\arraybackslash}m{2.4cm}
     >{\centering\arraybackslash}m{1.2cm}}

\toprule[1.5pt]
     \makecell[c]{Domain} & Variants & Dataset & Data Time Range & Data Granularity & Reference \\
    \midrule[0.8pt]
     & 21 & the Scada wind farm in Turkey & 2018/1/1-2018/12/29 & 10m & \citep{lin2021wind} \\

     & - & Global horizontal solar radiation data & 1998/1/1-2007/12/1 & 1h & \citep{sorkun2017time} \\

     \makecell[c]{Energy} & - & Rooftop PV plant & 2015/1/1-2016/12/31 & 30m & \citep{torres2019deep} \\

     & 9 & UCI household electric power consumption & 2006/12-2010/11 & 1m & \citep{bu2020time} \\

     & - & Spanish electricity demand & 2014/01/02-2019/11/01 & 10m & \citep{lara2020temporal} \\

     & - & Electric Vehicles Power Consumption & 2015/3/2-2016/5/31 & 1h & \citep{lara2020temporal} \\
    \midrule[0.8pt]
     & - & CDC ILI data & 2010-2018 & 1d & \citep{wu2020deep} \\

     & 45 & DEAP & - & 1 interval & \citep{koelstra2011deap} \\

     \raisebox{-3.0ex}{\makecell[c]{Healthcare}} & 9 & Turkish COVID-19 data & 2020/3/27-2020/6/11 & 1d & \citep{koc2022forecasting} \\

     & 9 & COVID-19 dataset of Orissa state & 2020/1/30-2020/6/11 & 1d & \citep{dash2021deep} \\

    \midrule[0.8pt]
    
     & 207 & METR-LA & 2012/3/1-2012/6/30 & 5m & \citep{cai2020traffic} \\

     \makecell[c]{Transportation} & 325 & PeMS-BAY & 2017/1/1-2017/5/31 & 5m & \citep{cai2020traffic} \\

     & - & BJER4 & 2014/7/1-2014/8/31 & 5m & \citep{yu2017spatio} \\
    
    \midrule[0.8pt]
     & 6 & Daily data of Shenzhen & from 2015 & - & \citep{chen2022daily} \\

     \makecell[c]{Meteorology} & - & CHIRPS & 1981-2015 & - & \citep{funk2015climate} \\

     & - & WeatherBench & - & - & \citep{rasp2020weatherbench} \\
    \midrule[0.8pt]
     & 5 & S\&P500 & 1997/1/1-2016/12/1 & 1d & \citep{lee2020threshold} \\

     \makecell[c]{Economics} & 13 & NSE stocks data & 1996/1/1-2015/6/30 & 1d & \citep{hiransha2018nse} \\

     & 6 & NYSE stock data & 2011/1/3-2016/12/30 & 1m & \citep{hiransha2018nse} \\

\bottomrule[1.5pt]

  \end{tabular}
\end{table}

\subsubsection{M3} This dataset \footnote{https://forecasters.org/resources/time-series-data/} comprises yearly, quarterly, monthly, daily, and other time series. To ensure the development of accurate forecasting models, minimum observation thresholds were established: 14 for yearly series, 16 for quarterly series, 48 for monthly series, and 60 for other series. Time series within the domains of micro, industry, macro, finance, demographic, and others were included.
\subsubsection{M4} The M4 dataset \citep{makridakis2018m4} encompasses 100,000 real-life series in diverse domains, including micro, industry, macro, finance, demographic, and others.
\subsubsection{M5} Covering stores in three US States (California, Texas, and Wisconsin), this dataset \footnote{https://mofc.unic.ac.cy/m5-competition/} includes item-level, department, product categories, and store details. It incorporates explanatory variables such as price, promotions, day of the week, and special events. Alongside time series data, it incorporates additional explanatory variables (e.g., Super Bowl, Valentine’s Day, and Orthodox Easter) influencing sales, enhancing forecasting accuracy.
\subsubsection{M6} The dataset \footnote{https://mofc.unic.ac.cy/} comprises two categories of assets: one selected from the Standard \& Poor's 500 Index, consisting of 50 stocks, and the other comprising 50 Exchange-Traded Funds (ETFs) from various international exchanges. The focus of the M6 competition lies in forecasting the returns and risks associated with these stocks, along with investment decisions made based on the aforementioned predictions.

\end{document}